\documentclass[12pt]{article}
\usepackage{soul}
\usepackage{amssymb}
\usepackage{lineno}
\usepackage{graphicx}
\usepackage{amsmath}
\usepackage{subfigure}
\usepackage{color}
\usepackage{url}
\usepackage{fancybox}
\usepackage{soul}
\usepackage{natbib}
\bibpunct{(}{)}{;}{a}{,}{,}

\begin{document}

\newcommand{\unorm}[1]{\|#1\|}
\newcommand{\unorms}[1]{\unorm{#1}^2}
\newcommand{\calX}{{\mathcal{X}}}
\newcommand{\calY}{{\mathcal{Y}}}
\newcommand{\boldtheta}{{\boldsymbol{\theta}}}
\newcommand{\boldalpha}{{\boldsymbol{\alpha}}}
\newcommand{\boldHh}{{\widehat{\boldH}}}
\newcommand{\boldH}{{\boldsymbol{H}}}
\newcommand{\boldA}{{\boldsymbol{A}}}
\newcommand{\boldS}{{\boldsymbol{S}}}
\newcommand{\boldK}{{\boldsymbol{K}}}
\newcommand{\boldJ}{{\boldsymbol{J}}}
\newcommand{\boldT}{{\boldsymbol{T}}}
\newcommand{\boldu}{{\boldsymbol{u}}}
\newcommand{\boldv}{{\boldsymbol{v}}}
\newcommand{\boldk}{{\boldsymbol{k}}}
\newcommand{\boldb}{{\boldsymbol{b}}}
\newcommand{\boldDelta}{{\boldsymbol{\Delta}}}
\newcommand{\nnu}{\nsample}
\newcommand{\nsample}{n}
\newcommand{\subsetr}{\boldsymbol{r}}
\newcommand{\boldthetah}{{\widehat{\boldtheta}}}
\newcommand{\argmin}{\mathop{\mathrm{argmin\,}}}
\newcommand{\mathbbR}{\mathbb{R}}
\newcommand{\numparams}{n}
\newcommand{\boldhh}{{\widehat{\boldh}}}
\newcommand{\boldh}{{\boldsymbol{h}}}
\newcommand{\Hh}{{\widehat{H}}}
\newcommand{\boldxnu}{\boldY}
\newcommand{\boldx}{{\boldsymbol{x}}}
\newcommand{\boldg}{{\boldsymbol{g}}}
\newcommand{\nde}{\nsample'}
\newcommand{\boldxde}{\boldY'}
\newcommand{\boldY}{{\boldsymbol{Y}}}
\newcommand{\boldy}{{\boldsymbol{y}}}
\newcommand{\boldYnu}{{\boldsymbol{Y}}}
\newcommand{\boldYde}{{\boldsymbol{Y}}}
\newcommand{\hh}{{\widehat{h}}}
\newcommand{\boldI}{{\boldsymbol{I}}}
\newcommand{\PE}{{\widehat{PE}}}
\newcommand{\ratioh}{\widehat{\ratiosymbol}}
\newcommand{\ratiosymbol}{r}
\newcommand{\ratiomodel}{g}
\newcommand{\thetah}{{\widehat{\theta}}}
\newcommand{\mathbbE}{\mathbb{E}}
\newcommand{\pnu}{p_\mathrm{te}}
\newcommand{\pde}{p_\mathrm{rf}}
\newcommand{\refsection}{\boldS_\mathrm{rf}}
\newcommand{\tesection}{\boldS_\mathrm{te}}
\newcommand{\refY}{\boldY_\mathrm{rf}}
\newcommand{\teY}{\boldY_\mathrm{te}}
\newcommand{\nseg}{n}
\newcommand{\argmax}{\mathop{\mathrm{argmax\,}}}
\providecommand{\e}[1]{\ensuremath{\times 10^{#1}}}
\def\ratio{r}
\def\relratio{{\ratio}_{\alpha}}

\providecommand{\songliu}[1]{\colorbox{yellow}{#1\\}}

\renewcommand{\subfigtopskip}{0mm}
\renewcommand{\subfigbottomskip}{0mm}
\renewcommand{\subfigcapskip}{0mm}

\setcounter{topnumber}{2}
\def\topfraction{1}
\setcounter{bottomnumber}{1}
\def\bottomfraction{1}
\setcounter{totalnumber}{10}
\def\textfraction{0}
\setcounter{dbltopnumber}{3}
\def\dbltopfraction{1}

\title{\vspace*{-33mm}
\begin{flushleft}
  \normalsize
  \sl
   \textit{Neural Networks, to appear.}
\end{flushleft}
\vspace*{10mm}
Change-Point Detection in Time-Series Data \\by Relative Density-Ratio Estimation 
}

\author{
Song Liu\\
Tokyo Institute of Technology\\
2-12-1 O-okayama, Meguro-ku, Tokyo 152-8552, Japan.\\
song@sg.cs.titech.ac.jp\\[2mm]
Makoto Yamada\\
NTT Communication Science Laboratories\\
2-4, Hikaridai, Seika-cho, Kyoto 619-0237, Japan.\\
yamada.makoto@lab.ntt.co.jp\\[2mm]
Nigel Collier\\
National Institute of Informatics (NII)\\
2-1-2 Hitotsubashi, Chiyoda-ku, Tokyo 101-8430, Japan.\\
European Bioinformatics Institute (EBI)\\
Wellcome Trust Genome Campus, Hinxton, Cambridge CB10 1SD, UK.\\
collier@nii.ac.jp\\[2mm]
Masashi Sugiyama\\
Tokyo Institute of Technology\\
2-12-1 O-okayama, Meguro-ku, Tokyo 152-8552, Japan.\\
sugi@cs.titech.ac.jp
\ \ http://sugiyama-www.cs.titech.ac.jp/\~{}sugi
}

\date{}

\maketitle


\begin{abstract} \noindent
The objective of change-point detection is to discover abrupt property
changes lying behind time-series data. In this paper, we present a
novel statistical change-point detection algorithm based on
non-parametric divergence estimation between time-series samples from
two retrospective
segments. Our method uses the relative Pearson divergence as a
divergence measure, and it is accurately and efficiently estimated by
a method of direct density-ratio estimation. Through experiments
on artificial and real-world datasets
including human-activity sensing, speech, and Twitter messages,
we demonstrate the usefulness of the proposed method.
\end{abstract}


\newpage

\section{Introduction}
Detecting abrupt changes in time-series data,
called \emph{change-point detection},
has attracted researchers in the statistics and data mining communities
for decades
\citep{Detection_Abrupt_Changes,book:Gustafsson:2000,book:Brodsky+Darkhovsky:1993}.
Depending on the delay of detection, change-point detection methods
can be classified into two categories:
\emph{Real-time detection} \citep{Bayisan_Online,Baysian_Online_2,Baysian_Online_3}
and
\emph{retrospective detection} \citep{Detection_Abrupt_Changes,AR_JOUR,SST_PAPER}.

Real-time change-point detection targets
applications that require immediate responses
such as robot control.
On the other hand,
although retrospective change-point detection
requires longer reaction periods,
it tends to give more robust and accurate detection.
Retrospective change-point detection
accommodates various applications that allow certain delays, for example,
climate change detection \citep{REVIEW_CLIMATE},
genetic time-series analysis \citep{CPDT_GENE},
signal segmentation \citep{Detection_Abrupt_Changes},
and intrusion detection in computer networks \citep{Intrusion_Detection}.
In this paper, we focus on the retrospective change-point detection scenario
and propose a novel non-parametric method.

Having been studied for decades, some pioneer works demonstrated good
change-point detection performance by comparing the probability distributions
of time-series samples over past and present intervals \citep{Detection_Abrupt_Changes}.
As both the intervals move forward, a typical strategy is to issue an alarm for
a change point when the two distributions are becoming significantly different.
Various change-point detection methods follow this strategy,
for example, the \emph{cumulative sum} \citep{Detection_Abrupt_Changes},
the \emph{generalized likelihood-ratio method} \citep{GenLR},
and
the \emph{change finder} \citep{AR_JOUR}.
Such a strategy has also been employed
in novelty detection \citep{LIKELIHOOD_ESTIMATION}
and outlier detection \citep{KAIS:Hido+etal:2011}.

Another group of methods that have attracted high popularity in recent years
is the \emph{subspace} methods \citep{SST_PAPER,SST_PAPER0,SST2_PAPER,SI_PAPER}.
By using a pre-designed time-series model, a subspace is discovered by
principal component analysis from trajectories in past
and present intervals, and their dissimilarity
is measured by the distance between the subspaces.
One of the major approaches is called
\textit{subspace identification}, which compares the subspaces spanned
by the columns of an \emph{extended observability matrix} generated by a state-space
model with system noise \citep{SI_PAPER}. Recent efforts along
this line of research have led to a computationally efficient algorithm based on
\emph{Krylov subspace learning} \citep{SST2_PAPER} and a successful application of
detecting climate change in south Kenya \citep{SST_CLIMATE_PAPER}.

The methods explained above
rely on pre-designed parametric models, such as
underlying probability distributions \citep{Detection_Abrupt_Changes, GenLR}, auto-regressive models \citep{AR_JOUR}, and state-space
models \citep{SST_PAPER,SST_PAPER0,SST2_PAPER,SI_PAPER}, for tracking specific statistics such as the mean, the variance, and the spectrum. As alternatives, non-parametric methods such as \emph{kernel density estimation} \citep{Inbook:Csorgo+Horvath:1988,book:Brodsky+Darkhovsky:1993} are designed with no particular parametric assumption. 
However, they tend to be less accurate
in high-dimensional problems because of the so-called \emph{curse of dimensionality}
\citep{book:Bellman:1961,book:Vapnik:1998}.

\begin{figure}[t]
\centering
\includegraphics[trim=0mm 5mm 0mm 5mm, clip,width=0.6\textwidth,clip]{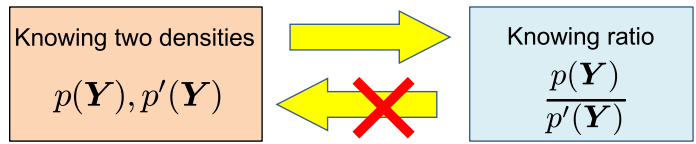}
\vspace*{-2mm}
\caption{Rationale of direct density-ratio estimation.}
\label{fig:rationale}
\end{figure}

To overcome this difficulty, a new strategy was introduced recently,
which estimates the \emph{ratio} of
probability densities directly without going through density estimation \citep{book:Sugiyama+etal:2012}.
The rationale of this density-ratio estimation idea is that
knowing the two densities implies knowing the density ratio,
but not vice versa;
knowing the ratio does not necessarily imply knowing the two densities
because such decomposition is not unique (Figure~\ref{fig:rationale}).
Thus, direct density-ratio estimation is substantially
easier than density estimation \citep{book:Sugiyama+etal:2012}.
Following this idea, methods of direct density-ratio estimation
have been developed \citep{ULSIF_STABILITY2},
e.g., \emph{kernel mean matching} \citep{Inbook:Gretton+etal:2009},
the \emph{logistic-regression method} \citep{ICML:Bickel+etal:2007},
and the \emph{Kullback-Leibler importance estimation procedure} (KLIEP)
\citep{KLIEP_PAPER}.
In the context of change-point detection, KLIEP was reported to outperform
other approaches \citep{CPDT_JOUR}
such as the \emph{one-class support vector machine} \citep{ONE_SVM,ONESVM_PAPER}
and \emph{singular-spectrum analysis} \citep{SST_PAPER0}.
Thus, change-point detection based on direct density-ratio estimation
is promising.

The goal of this paper is to further advance this line of research.
More specifically, our contributions in this paper are two folds.
The first contribution is to apply a recently-proposed
density-ratio estimation method called
the \emph{unconstrained least-squares importance fitting} (uLSIF) \citep{ULSIF}
to change-point detection.
The basic idea of uLSIF is to directly learn the density-ratio function
in the least-squares fitting framework.
Notable advantages of uLSIF are that its solution can be
computed analytically \citep{ULSIF},
it achieves the optimal non-parametric convergence rate \citep{ML:Kanamori+etal:2011},
it has the optimal numerical stability \citep{ULSIF_STABILITY},
and it has higher robustness than KLIEP \citep{ULSIF_STABILITY2}.
Through experiments on a range of datasets, we demonstrate the superior detection accuracy of
the uLSIF-based change-point detection method.

The second contribution of this paper
is to further improve the uLSIF-based change-point detection method
by employing a state-of-the-art extension of uLSIF called \emph{relative uLSIF} (RuLSIF)
\citep{RULSIF_NIPS}.
A potential weakness of the density-ratio based approach is that
density ratios can be unbounded (i.e., they can be infinity) if the denominator density
is not well-defined.
The basic idea of RuLSIF is to consider \emph{relative density ratios},
which are smoother and always bounded from above.
Theoretically, it was proved that RuLSIF possesses a superior
non-parametric convergence property than plain uLSIF \citep{RULSIF_NIPS}, implying
that RuLSIF gives an even better estimate from a small number of samples.
We experimentally demonstrate that our RuLSIF-based change-point detection method
compares favorably with other approaches.



The rest of this paper is structured as follows:
In Section~\ref{sec.problem.form}, we formulate our change-point detection problem.
In Section~\ref{sec:density-ratio}, we describe our proposed change-point detection
algorithms based on uLSIF and RuLSIF, together with the review of the KLIEP-based method.
In Section~\ref{sec:experiment}, we report experimental results
on various artificial and real-world datasets
including human-activity sensing, speech, and Twitter messages from February 2010 to October 2010.
Finally, in Section~\ref{sec:conclusion},
conclusions together with future perspectives are stated.

\begin{figure}[t]
\centering
\includegraphics[width=0.48\textwidth,clip]{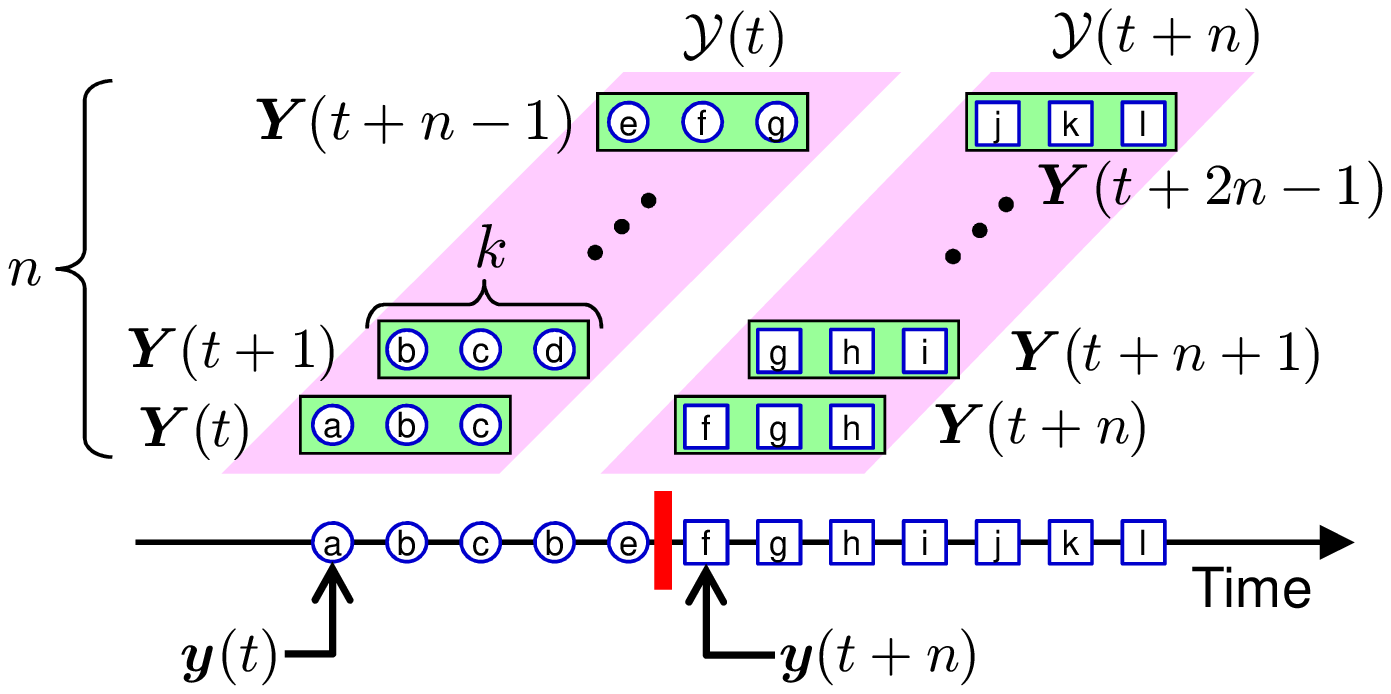}
\includegraphics[width=0.48\textwidth,clip]{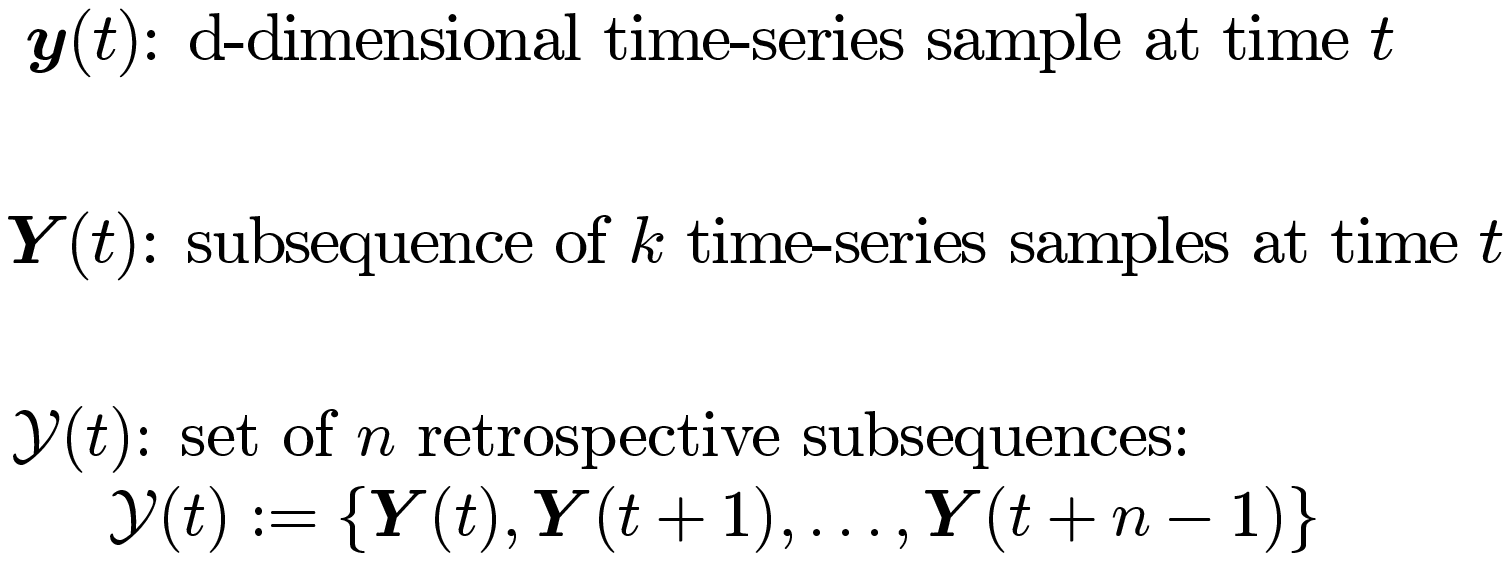}
\vspace*{-2mm}
\caption{An illustrative example of notations on one-dimensional time-series data. }
\label{fig:illus_sec1}
\end{figure}

\section{Problem Formulation} \label{sec.problem.form}
In this section, we formulate our change-point detection problem.



Let $\boldy(t)\in \mathbb{R}^d$
be a $d$-dimensional time-series sample at time $t$.
Let
\[
\boldY(t) := [ \boldy(t)^\top, \boldy(t+1)^\top, \ldots,\boldy(t+k-1)^\top]^\top
\in \mathbb{R}^{dk}
\]
be a ``subsequence''\footnote{In fact, only in the case of one-dimensional time-series, $\boldY(t)$ is a subsequence. For higher-dimensional time-series, $\boldY(t)$ concatenates the subsequences of all dimensions into a one-dimensional vector. } of time series at time $t$ with length $k$,
where ${}^\top$ represents the transpose.
Following the previous work  \citep{CPDT_JOUR}, we treat the subsequence $\boldY(t)$ as a sample,
instead of a single $d$-dimensional time-series sample $\boldy(t)$,
by which time-dependent information can be incorporated naturally
(see Figure~\ref{fig:illus_sec1}).
Let $\calY(t)$ be a set of $n$ retrospective subsequence samples
starting at time $t$:
\begin{align*}
&\calY(t) := \{{\boldY(t)},\boldY(t+1),\hdots,\boldY(t+n-1)\}.
\end{align*}
Note that $[{\boldY(t)},\boldY(t+1),\hdots,\boldY(t+n-1)]\in \mathbb{R}^{dk\times n}$
forms a \textit{Hankel matrix}
and plays a key role in change-point detection
based on subspace learning \citep{SST_PAPER,SI_PAPER}.



For change-point detection, let us consider two consecutive segments
$\calY(t)$ and $\calY(t+n)$.
Our strategy is to compute a certain dissimilarity measure
between $\calY(t)$ and $\calY(t+n)$,
and use it as the plausibility of change points.
More specifically, the higher the dissimilarity measure is,
the more likely the point is a change point\footnote{
Another possible formulation is to compare
distributions of samples in $\calY(t)$ and $\calY(t+n)$
in the framework of \emph{hypothesis testing} \citep{book:Henkel:1976}.
Although this gives a useful threshold to determine whether
a point is a change point, computing the $p$-value is often time consuming,
particularly in a non-parametric setup \citep{book:Efron+Tibshirani:1993}.
For this reason, we do not take the hypothesis testing approach in this paper,
although it is methodologically straightforward to extend the proposed approach
to the hypothesis testing framework.
}.

Now the problems that need to be addressed are
what kind of dissimilarity measure we should use
and how we estimate it from data.
We will discuss these issues in the next section.

\section{Change-Point Detection via Density-Ratio Estimation}\label{sec:density-ratio}
In this section, we first define our dissimilarity measure,
and then show methods for estimating the dissimilarity measure.

\subsection{Divergence-Based Dissimilarity Measure and Density-Ratio Estimation}
\label{sec:div_based_diss}
In this paper, we use a dissimilarity measure of the following form:
\begin{align}
  D(P_t \| P_{t+n}) + D(P_{t+n} \| P_{t}),
\label{dissimilarity}
\end{align}
where $P_t$ and $P_{t+n}$ are probability distributions
of samples in $\calY(t)$ and $\calY(t+n)$, respectively.
$D(P\|P')$ denotes the \emph{$f$-divergence} \citep{JRSS-B:Ali+Silvey:1966,SSM-Hungary:Csiszar:1967}:
\begin{align}
\label{eq.f-divergence}
  D(P\|P'):=\int p'(\boldY)f\left(\frac{p(\boldY)}{p'(\boldY)}\right)\mathrm{d}\boldY,
\end{align}
where $f$ is a convex function such that $f(1)=0$,
and $p(\boldY)$ and $p'(\boldY)$ are probability density functions of $P$ and $P'$, respectively.
We assume that $p(\boldY)$ and $p'(\boldY)$ are strictly positive.
Since the $f$-divergence is asymmetric (i.e., $D(P\|P')\neq D(P'\|P)$),
we symmetrize it in our dissimilarity measure \eqref{dissimilarity} for all divergence-based methods\footnote{In the previous work \citep{CPDT_JOUR}, the asymmetric dissimilarity measure $D(P_t||P_{t+n})$ was used. As we numerically illustrate in Section~\ref{sec:experiment}, the use of the symmetrized divergence contributes highly to improving the performance.
For this reason, we decided to use the symmetrized dissimilarity measure \eqref{dissimilarity}.}.

The $f$-divergence includes various popular divergences
such as the \emph{Kullback-Leibler (KL) divergence} by $f(t)=t\log t$
\citep{Annals-Math-Stat:Kullback+Leibler:1951}
and
the \emph{Pearson (PE) divergence} by $f(t)=\frac{1}{2}(t-1)^2$
\citep{PE_DIV}:
\begin{align}
\mathrm{KL}(P\|P')&:=\int p(\boldY)\log\left(\frac{p(\boldY)}{p'(\boldY)}\right)
\mathrm{d}\boldY,\label{KL}\\
\mathrm{PE}(P\|P')&:=\frac{1}{2} \int p'(\boldY)
\left(\frac{p(\boldY)}{p'(\boldY)}-1\right)^2 \mathrm{d}\boldY.\label{PE}
\end{align}

Since the probability densities $p(\boldY)$ and $p'(\boldY)$ are unknown in practice,
we cannot directly compute the $f$-divergence (and thus the dissimilarity measure).
A naive way to cope with this problem is to perform density estimation and plug the estimated
densities $\widehat{p}(\boldY)$ and $\widehat{p}'(\boldY)$ in the
definition of the $f$-divergence.
However, density estimation is known to be a hard problem \citep{book:Vapnik:1998},
and thus such a plug-in approach is not reliable in practice.

Recently, a novel method of divergence approximation
based on \emph{direct density-ratio estimation} was explored
\citep{KLIEP_PAPER,IEEE-IT:Nguyen+etal:2010,ULSIF}.
The basic idea of direct density-ratio estimation is to learn
the density-ratio function $\frac{p(\boldY)}{p'(\boldY)}$ without going through
separate density estimation of $p(\boldY)$ and $p'(\boldY)$.
An intuitive rationale of direct density-ratio estimation
is that knowing the two densities $p(\boldY)$ and $p'(\boldY)$
means knowing their ratio, but not vice versa;
knowing the ratio $\frac{p(\boldY)}{p'(\boldY)}$ does not necessarily
mean knowing the two densities $p(\boldY)$ and $p'(\boldY)$
because such decomposition is not unique
(see Figure~\ref{fig:rationale}).
This implies that estimating the density ratio is substantially easier
than estimating the densities,
and thus directly estimating the density ratio would be more promising\footnote{
Vladimir Vapnik advocated in his seminal book \citep{book:Vapnik:1998}
that one should avoid solving a more difficult problem as an intermediate step.
The \emph{support vector machine} \citep{mach:Cortes+Vapnik:1995}
is a representative example that demonstrates the usefulness of this principle:
It avoids solving a more general problem of estimating data generating probability distributions,
and only learns a decision boundary that is sufficient for pattern recognition.
The idea of direct density-ratio estimation also follows Vapnik's principle.
}
\citep{book:Sugiyama+etal:2012}.

In the rest of this section, we review three methods of directly
estimating the  density ratio $\frac{p(\boldY)}{p'(\boldY)}$
from samples $\{\boldY_i\}_{i=1}^{\nseg}$ and $\{\boldY'_j\}_{j=1}^{\nseg}$
drawn from $p(\boldY)$ and $p'(\boldY)$:
The \emph{KL importance estimation procedure} (KLIEP)
\citep{KLIEP_PAPER} in Section~\ref{sec:KLIEP},
\emph{unconstrained least-squares importance fitting} (uLSIF)
\citep{ULSIF} in Section~\ref{sec.ulsif},
and \emph{relative uLSIF (RuLSIF)} \citep{RULSIF_NIPS} in Section~\ref{sec:RuLSIF}.

\subsection{KLIEP}\label{sec:KLIEP}

KLIEP \citep{KLIEP_PAPER}
is a direct density-ratio estimation algorithm
that is suitable for estimating the KL divergence.

\subsubsection{Density-Ratio Model}\label{subsec:KLIEP-model}

Let us model the density ratio $\frac{p(\boldY)}{p'(\boldY)}$
by the following kernel model:
\begin{align}\label{eq.density.ratio.model}
  g(\boldY;\boldtheta):=\sum_{\ell=1}^{n}
  \theta_\ell K(\boldY,\boldY_\ell),
\end{align}
where $\boldtheta:=(\theta_1,\ldots,\theta_{n})^\top$
are parameters to be learned from data samples,
and $K(\boldY,\boldY')$ is a kernel basis function.
In practice, we use the Gaussian kernel:
\begin{align*}
  K(\boldY,\boldY')=
  \exp\left(-\frac{\|\boldY-\boldY'\|^2}{2\sigma^2}\right),
\end{align*}
where $\sigma$ ($>0$) is the kernel width.
In all our experiments, the kernel width $\sigma$ is determined based on cross-validation.

\subsubsection{Learning Algorithm}
The parameters $\boldtheta$ in the model $g(\boldY; \boldtheta)$ are determined so that
the KL divergence from $p(\boldY)$ to $g(\boldY; \boldtheta)p'(\boldY)$ is minimized:
\begin{align*}
\mathrm{KL}
 &= \int p(\boldY) \log\left(\frac{p(\boldY)}{p'(\boldY)g(\boldY;\boldtheta)}\right) \; \mathrm{d}\boldY\\
& = \int p(\boldY) \log\left(\frac{p(\boldY)}{p'(\boldY)}\right) \; \mathrm{d}\boldY - \int p(\boldY) \log\left(g(\boldY;\boldtheta)\right) \; \mathrm{d}\boldY
\end{align*}
After ignoring the first term which is irrelevant to $g(\boldY;\boldtheta)$ and approximating the second term with the empirical estimates, the KLIEP optimization problem is given as follows:
\begin{align*}
\max_{\boldtheta}&\; \frac{1}{\nsample}\sum^{\nsample}_{i=1}\log\left(\sum_{\ell = 1}^{\nsample}{\theta_\ell K(\boldY_i,\boldY_\ell)}\right),\\
\text{s.t.}&\;\frac{1}{\nsample}\sum_{j=1}^{\nsample}\sum_{\ell=1}^{\nsample}
\theta_{\ell}K(\boldY'_j,\boldY_\ell)= 1
\text{ and } \theta_1,\ldots,\theta_n \ge 0.
\end{align*}
The equality constraint is for the normalization purpose
because $g(\boldY; \boldtheta)p'(\boldY)$ should be a probability density function.
The inequality constraint comes from the non-negativity of
the density-ratio function.
Since this is a convex optimization problem,
the unique global optimal solution $\widehat{\boldtheta}$ can be simply obtained,
for example, by a gradient-projection iteration.
Finally, a density-ratio estimator is given as
\begin{align*}
  \widehat{g}(\boldY)=\sum_{\ell=1}^{\nnu} \thetah_\ell K(\boldY,\boldY_\ell).
\end{align*}

KLIEP was shown to achieve the optimal non-parametric convergence rate
\citep{KLIEP_PAPER,IEEE-IT:Nguyen+etal:2010}.

\subsubsection{Change-Point Detection by KLIEP}
Given a density-ratio estimator $\widehat{g}(\boldY)$,
an approximator of the KL divergence is given as
\begin{align*}
  \widehat{\mathrm{KL}}:=\frac{1}{n}\sum_{i=1}^n \log\widehat{g}(\boldY_i).
\end{align*}

In the previous work \citep{CPDT_JOUR},
this KLIEP-based KL-divergence estimator was applied to change-point detection
and demonstrated to be promising in experiments.

\subsection{uLSIF}\label{sec.ulsif}

Recently, another direct density-ratio estimator
called uLSIF was proposed \citep{ULSIF,ML:Kanamori+etal:2011},
which is suitable for estimating the PE divergence.

\subsubsection{Learning Algorithm}
\label{sec:ulsif.learning}
In uLSIF, the same density-ratio model as KLIEP is used (see Section~\ref{subsec:KLIEP-model}).
However, its training criterion is different; the density-ratio model
is fitted to the true density-ratio under the squared loss.
More specifically, the parameter $\boldtheta$ in the model $g(\boldY;\boldtheta)$ is determined so that the following squared loss $J(\boldY)$ is minimized:
\begin{align*}
J(\boldY) &= \frac{1}{2}\int \left(\frac{p(\boldY)}{p'(\boldY)} - g(\boldY; \boldtheta)\right)^2
p'(\boldY) \; \mathrm{d}\boldY\\
&= \frac{1}{2}\int \left(\frac{p(\boldY)}{p'(\boldY)}\right)^2 p'(\boldY) \; \mathrm{d}\boldY - \int p(\boldY)g(\boldY;\boldtheta) \; \mathrm{d}\boldY + \frac{1}{2} \int g(\boldY;\boldtheta)^2 p'(\boldY) \;\mathrm{d}\boldY.
\end{align*}
Since the first term is a constant, we focus on the last two terms. By substituting $g(\boldY;\boldtheta)$ with our model stated in \eqref{eq.density.ratio.model} and approximating the integrals by the empirical averages, the uLSIF optimization problem is given as follows:

\begin{align}
  \min_{\boldtheta\in\mathbbR^{\numparams}}
  \left[\frac{1}{2}\boldtheta^\top\boldHh\boldtheta-\boldhh^\top\boldtheta
  +\frac{\lambda}{2}\boldtheta^\top\boldtheta\right],
  \label{uLSIF-optimization-empirical}
\end{align}
where the penalty term
$\frac{\lambda}{2}\boldtheta^\top\boldtheta$ is included for a regularization purpose.
$\lambda$ $(\ge0)$ denotes the regularization parameter,
which is chosen by cross-validation \citep{KLIEP_PAPER}.
$\boldHh$ is the $n\times n$ matrix with the $(\ell,\ell')$-th element given by
\begin{align}\label{Hh}
  \widehat{H}_{\ell,\ell'}:= &
  \frac{1}{n}\sum_{j=1}^{n} K(\boldY'_j,\boldYnu_\ell)K(\boldY'_j,\boldYnu_{\ell'}).
\end{align}
$\boldhh$ is the $n$-dimensional vector with the $\ell$-th element given by
\begin{align*}
  \hh_{\ell}:=\frac{1}{n}\sum_{i=1}^{n} K(\boldY_i,\boldY_\ell).
\end{align*}

It is easy to confirm that the solution $\boldthetah$
of \eqref{uLSIF-optimization-empirical} can be analytically obtained as
\begin{equation}
\label{sol}
  \boldthetah=(\boldHh+\lambda\boldI_{\nnu})^{-1}\boldhh,
\end{equation}
where $\boldI_{\nnu}$ denotes the $\nnu$-dimensional identity matrix.
Finally, a density-ratio estimator is given as
\begin{align*}
  \widehat{g}(\boldY)=\sum_{\ell=1}^{\nnu} \thetah_\ell K(\boldY,\boldY_\ell).
\end{align*}

\subsubsection{Change-Point Detection by uLSIF}
Given a density-ratio estimator $\widehat{g}(\boldY)$,
an approximator of the PE divergence can be constructed as
\begin{align*}
  \widehat{\mathrm{PE}}:=-\frac{1}{2n}\sum_{j=1}^n \widehat{g}(\boldY'_j)^2
  +\frac{1}{n}\sum_{i=1}^n \widehat{g}(\boldY_i)-\frac{1}{2}.
\end{align*}
This approximator is derived from the following expression of the PE divergence \citep{NN:Sugiyama+etal:2010, NN:Sugiyama+etal:2011a}:

\begin{align}
\label{eq.fen-dual}
\mathrm{PE}(P\|P')&=-\frac{1}{2} \int \left(\frac{p(\boldY)}{p'(\boldY)}\right)^2
p'(\boldY)\mathrm{d}\boldY
+\int \left(\frac{p(\boldY)}{p'(\boldY)}\right)p(\boldY)\mathrm{d}\boldY
-\frac{1}{2}.
\end{align}

The first two terms of \eqref{eq.fen-dual} are actually the negative uLSIF optimization objective without regularization. This expression can also be obtained based on the fact that the $f$-divergence $D(P\|P')$ is lower-bounded via the \emph{Legendre-Fenchel convex duality} \citep{book:Rockafellar:1970} as follows \citep{Keziou03,Nguyen07}:
\begin{align}
\label{eq.lower-bound}
D(P||P') = \sup_{h}\left( \int p(\boldY)h(\boldY) \; \mathrm{d}\boldY - \int p'(\boldY) f^*(h(\boldY)) \; \mathrm{d}\boldY \right),
\end{align}
where $f^*$ is the convex conjugate of convex function $f$ defined at \eqref{eq.f-divergence}.
The PE divergence corresponds to $f(t) = \frac{1}{2}(t-1)^2$, for which
convex conjugate is given by $f^*(t^*) = \frac{{(t^*)}^2}{2} + t^*$. For $f(t) = \frac{1}{2}(t-1)^2$, the supremum can be achieved when $\frac{p(\boldY)}{p'(\boldY)} = h(\boldY) + 1$. Substituting $h(\boldY) = \frac{p(\boldY)}{p'(\boldY)}  - 1$ into \eqref{eq.lower-bound}, we can obtain \eqref{eq.fen-dual}.

uLSIF has some notable advantages: Its solution can be
computed analytically \citep{ULSIF} and it possesses the optimal non-parametric convergence rate \citep{ML:Kanamori+etal:2011}. Moreover,
it has the optimal numerical stability \citep{ULSIF_STABILITY},
and it is more robust than KLIEP \citep{ULSIF_STABILITY2}.
In Section~\ref{sec:experiment},
we will experimentally demonstrate that uLSIF-based change-point detection
compares favorably with the KLIEP-based method.

\subsection{RuLSIF} \label{sec:RuLSIF}

Depending on the condition of the denominator density $p'(\boldY)$,
the density-ratio value $\frac{p(\boldY)}{p'(\boldY)}$
can be unbounded (i.e., they can be infinity).
This is actually problematic because the non-parametric convergence rate of uLSIF is
governed by the ``sup''-norm of the true
density-ratio function: $\max_{\boldY}\frac{p(\boldY)}{p'(\boldY)}$.
To overcome this problem, \emph{relative density-ratio estimation}
was introduced \citep{RULSIF_NIPS}.

\subsubsection{Relative PE Divergence}
Let us consider the \emph{$\alpha$-relative PE-divergence} for $0\le\alpha<1$:
  \begin{align*}
    \mathrm{PE}_\alpha(P\|P')&:=\mathrm{PE}(P\|\alpha P+(1-\alpha)P')\\
  &\phantom{:}=\int  p'_\alpha(\boldY) \left(\frac{p(\boldY)}{p'_{\alpha}(\boldY)}-1\right)^2
  \mathrm{d}\boldY,
  \end{align*}
where $p'_\alpha(\boldY) = \alpha p(\boldY) + (1-\alpha) p'(\boldY)$ is the \emph{$\alpha$-mixture density}.
We refer to
\[
r_\alpha(\boldY) = \frac{p(\boldY)}{\alpha p(\boldY)+(1-\alpha)p'(\boldY)}
\]
as the \emph{$\alpha$-relative density-ratio}.
The $\alpha$-relative density-ratio is reduced to the plain density-ratio if $\alpha=0$,
and it tends to be ``smoother'' as $\alpha$ gets larger.
Indeed, one can confirm that the $\alpha$-relative density-ratio
is bounded above by $1/\alpha$ for $\alpha>0$,
even when the plain density-ratio $\frac{p(\boldY)}{p'(\boldY)}$ is unbounded.
This was proved to contribute to improving the estimation accuracy \citep{RULSIF_NIPS}.

As explained in Section \ref{sec:div_based_diss}, we use symmetrized divergence
\[
\mathrm{PE}_\alpha(P\|P') + \mathrm{PE}_\alpha(P'\|P)
\]
as a change-point score, where each term is estimated separately. 

\subsubsection{Learning Algorithm}
For approximating the $\alpha$-relative density ratio $r_\alpha(\boldY)$,
we still use the same kernel model $g(\boldY; \boldtheta)$ given by \eqref{eq.density.ratio.model}.
In the same way as the uLSIF method, the parameter $\boldtheta$ is learned by minimizing the squared loss between true and estimated relative ratios:
\begin{align*}
J(\boldY) &= \frac{1}{2}\int p'_\alpha(\boldY) \Big(r_\alpha(\boldY) - g(\boldY;\boldtheta)\Big)^2 \; \mathrm{d}\boldY \\
& = \frac{1}{2}\int p'_\alpha(\boldY) r^2_\alpha(\boldY)\; \mathrm{d}\boldY -  \int p(\boldY) r_\alpha(\boldY)g(\boldY; \boldtheta) \; \mathrm{d}\boldY
\\
&\phantom{=}+ \frac{\alpha}{2}\int p(\boldY)g(\boldY; \boldtheta)^2 \; \mathrm{d}\boldY
+ \frac{1-\alpha}{2}\int p'(\boldY)g(\boldY; \boldtheta)^2 \; \mathrm{d}\boldY.
\end{align*}

Again, by ignoring the constant and approximating the expectations by sample averages,
the $\alpha$-relative density-ratio can be learned
in the same way as the plain density-ratio.
Indeed, the optimization problem of a relative variant of uLSIF, called RuLSIF,
is given as the same form as uLSIF; the only difference is the definition of
the matrix $\boldHh$:
\begin{align*}
  \widehat{H}_{\ell,\ell'}&:=
  \frac{\alpha}{n}\sum_{i=1}^{n} K(\boldY_i,\boldY_\ell)K(\boldY_i,\boldY_{\ell'}) + \frac{(1-\alpha)}{n} \sum_{j=1}^{n}
  K(\boldY'_j,\boldY_\ell)K(\boldY'_j,\boldY_{\ell'}).
\end{align*}
Thus, the advantages of uLSIF regarding the analytic solution, numerical stability, and robustness are still maintained in RuLSIF.
Furthermore, RuLSIF possesses an even better non-parametric convergence property
than uLSIF \citep{RULSIF_NIPS}.

\subsubsection{Change-Point Detection by RuLSIF}
By using an estimator $\widehat{g}(\boldY)$ of
the $\alpha$-relative density-ratio,
the $\alpha$-relative PE divergence can be approximated as
\begin{align*}
  \widehat{\mathrm{PE}}_\alpha&:=
  -\frac{\alpha}{2n}\sum_{i=1}^n \widehat{g}(\boldY_i)^2
  -\frac{1-\alpha}{2n}\sum_{j=1}^n \widehat{g}(\boldY'_j)^2+\frac{1}{n}\sum_{i=1}^n \widehat{g}(\boldY_i)-\frac{1}{2}.
\end{align*}

In Section~\ref{sec:experiment},
we will experimentally demonstrate that the RuLSIF-based change-point detection
performs even better than the plain uLSIF-based method.

\section{Experiments}
\label{sec:experiment}

In this section, we experimentally investigate the performance of
the proposed and existing change-point detection methods on artificial and real-world datasets
including human-activity sensing, speech, and Twitter messages.
The MATLAB implementation of the proposed method is available at
\begin{center}
``\url{http://sugiyama-www.cs.titech.ac.jp/~song/change_detection/}''.
\end{center}

For all experiments, we fix the parameters at $n = 50$ and $k =  10$.
$\alpha$ in the RuLSIF-based method is fixed to $0.1$.
Sensitivity to different parameter choices
and more issues regarding algorithm-specific parameter tuning
will be discussed below.


\subsection{Artificial Datasets}
\label{sec.artifi}

\begin{figure}[t]
\centering
\includegraphics[width=.7\textwidth]{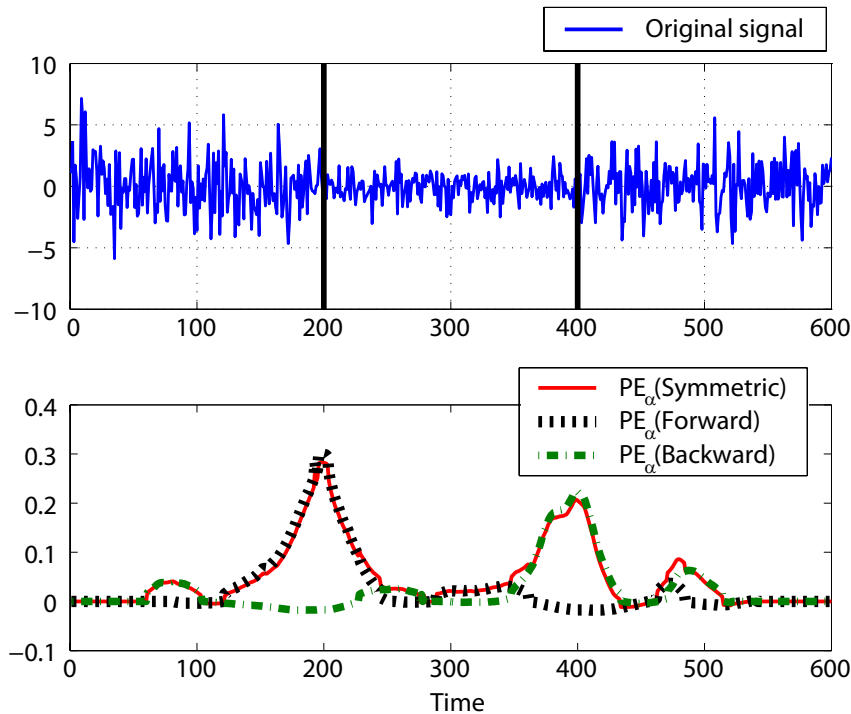}
\caption{(Top) The original signal (blue) is segmented into 3 sections with equal length. Samples are drawn from the normal distributions $\mathcal{N}(0,2^2)$, $\mathcal{N}(0,1^2)$, and $\mathcal{N}(0,2^2)$, respectively. (Bottom) Symmetric (red) and asymmetric (black and green) $\text{PE}_{\alpha}$ divergences. }
\label{symm_non_symm}
\end{figure}

As mentioned in Section~\ref{sec:div_based_diss}, we use the symmetrized  divergence for change-point detection. We first illustrate how symmetrization of the PE divergence affects the change-point detection performance.

The top graph in Figure~\ref{symm_non_symm} shows an artificial time-series signal that consists of three segments with equal length of 200. The samples are drawn from $\mathcal{N}(0,2^2)$, $\mathcal{N}(0,1^2)$, and $\mathcal{N}(0,2^2)$, respectively,
where $\mathcal{N}(\mu,\sigma^2)$ denotes the normal distribution with mean $\mu$ and variance $\sigma^2$.
 Thus, the variances change at time 200 and 400. In this experiment, we consider three types of divergence measures:
\begin{itemize}
	\item $\text{PE}_{\alpha}(\text{Symmetric}): \text{PE}_\alpha(P_t||P_{t+n}) + \text{PE}_\alpha(P_{t+n}||P_t)$,
	\item $\text{PE}_{\alpha}(\text{Forward}): \text{PE}_\alpha(P_t||P_{t+n})$,
	\item $\text{PE}_{\alpha}(\text{Backward}): \text{PE}_\alpha(P_{t+n}||P_{t})$.
\end{itemize}
Three divergences are compared in the bottom graph of Figure~\ref{symm_non_symm}.

As we can see from the graphs, $\text{PE}_\alpha(\text{Forward})$ detects the first change point successfully, but not the second one. On the other hand, $\text{PE}_\alpha(\text{Backward})$ behaves oppositely. This implies that combining forward and backward divergences can improve the overall change-point detection performance.
For this reason, we only use $\text{PE}_{\alpha}$(Symmetric) as the change-point score of the proposed method from here on.



Next, we illustrate the behavior of our proposed RuLSIF-based method,
and then compare its performance with the uLSIF-based and KLIEP-based methods. In our implementation, two sets of candidate parameters,
\begin{itemize}
\item $\sigma= 0.6d_\mathrm{med}$, $0.8d_\mathrm{med}$, $d_\mathrm{med}$, $1.2d_\mathrm{med}$,
  and $1.4d_\mathrm{med}$,
\item $\lambda=10^{-3}$, $10^{-2}$, $10^{-1}$, $10^{0}$, and $10^{1}$,
\end{itemize}
  are provided to the cross-validation procedure, where $d_\mathrm{med}$ denotes the median distance between samples. The best combination of these parameters is chosen by grid search via cross-validation. We use 5-fold cross-validation for all experiments.

We use the following 4 artificial time-series datasets that
contain manually inserted change-points:
\begin{itemize}
 \item \textbf{Dataset 1 (Jumping mean):}
   The following 1-dimensional
   auto-regressive model borrowed from \citet{AR_JOUR} is used
   to generate 5000 samples (i.e., $t = 1, \ldots, 5000$):
 \[y(t) = 0.6y(t-1) - 0.5y(t-2) + \epsilon_t,\]
 where $\epsilon_t$ is a Gaussian noise with mean $\mu$ and standard deviation $1.5$.
 The initial values are set as $y(1) = y(2)=0$.
 A change point is inserted at every $100$ time steps
 by setting the noise mean $\mu$ at time $t$ as
 \begin{equation*}
  \\ \mu_N =
  \begin{cases}
  0 & N=1,\\
  \mu_{N-1} + \frac{N}{16} & N=2,\ldots,49,
  \end{cases}
 \end{equation*}
where $N$ is a natural number such that $100(N-1)+1\le t\le100N$.

 \item \textbf{Dataset 2  (Scaling variance):}
   The same auto-regressive model as Dataset 1 is used,
   but a change point is inserted at every $100$ time steps
   by setting the noise standard deviation $\sigma$ at time $t$ as
 \begin{equation*}
 \sigma =
 \begin{cases}
   1&  N = 1,3,\ldots ,49,\\
   \ln(e+\frac{N}{4}) &  N = 2,4, \ldots ,48.\\
  \end{cases}
 \end{equation*}

 \item \textbf{Dataset 3 (Switching covariance):}
   2-dimensional samples of size 5000 are drawn from
   the origin-centered normal distribution,
   and a change point is inserted at every 100 time steps
   by setting the covariance matrix $\boldsymbol{\Sigma}$ at time $t$ as
  \begin{equation*}
    \boldsymbol{\Sigma} =
 \begin{cases}
  \begin{pmatrix}
   1 & -\frac{4}{5} - \frac{N-2}{500} \\
   -\frac{4}{5} - \frac{N-2}{500} & 1
  \end{pmatrix} &  N = 1,3, \ldots ,49,\\
\\[-3mm]
   \begin{pmatrix}
   1 & \frac{4}{5} + \frac{N-2}{500} \\
   \frac{4}{5} + \frac{N-2}{500} & 1
  \end{pmatrix}&  N = 2,4, \ldots ,48.
  \end{cases}
 \end{equation*}

\item \textbf{Dataset 4 (Changing frequency):}
  1-dimensional samples of size 5000 are generated as
 \[y(t) = \sin(\omega x)+ \epsilon_t,\]
 where $\epsilon_t$ is a origin-centered Gaussian noise with standard deviation $0.8$.
A change point is inserted at every 100 points by changing
  the frequency $\omega$ at time $t$ as
 \begin{equation*}
  \omega_N =
  \begin{cases}
  1 & N=1,\\
  \omega_{N-1} \ln(e + \frac{N}{2}) & N=2,\ldots,49.
  \end{cases}
 \end{equation*}
\end{itemize}

\begin{figure*}[p]
  \begin{minipage}[t]{0.6\textwidth}
\centering
\subfigure[Dataset1]{
 \includegraphics[width=\textwidth,height=.18\textheight,clip]{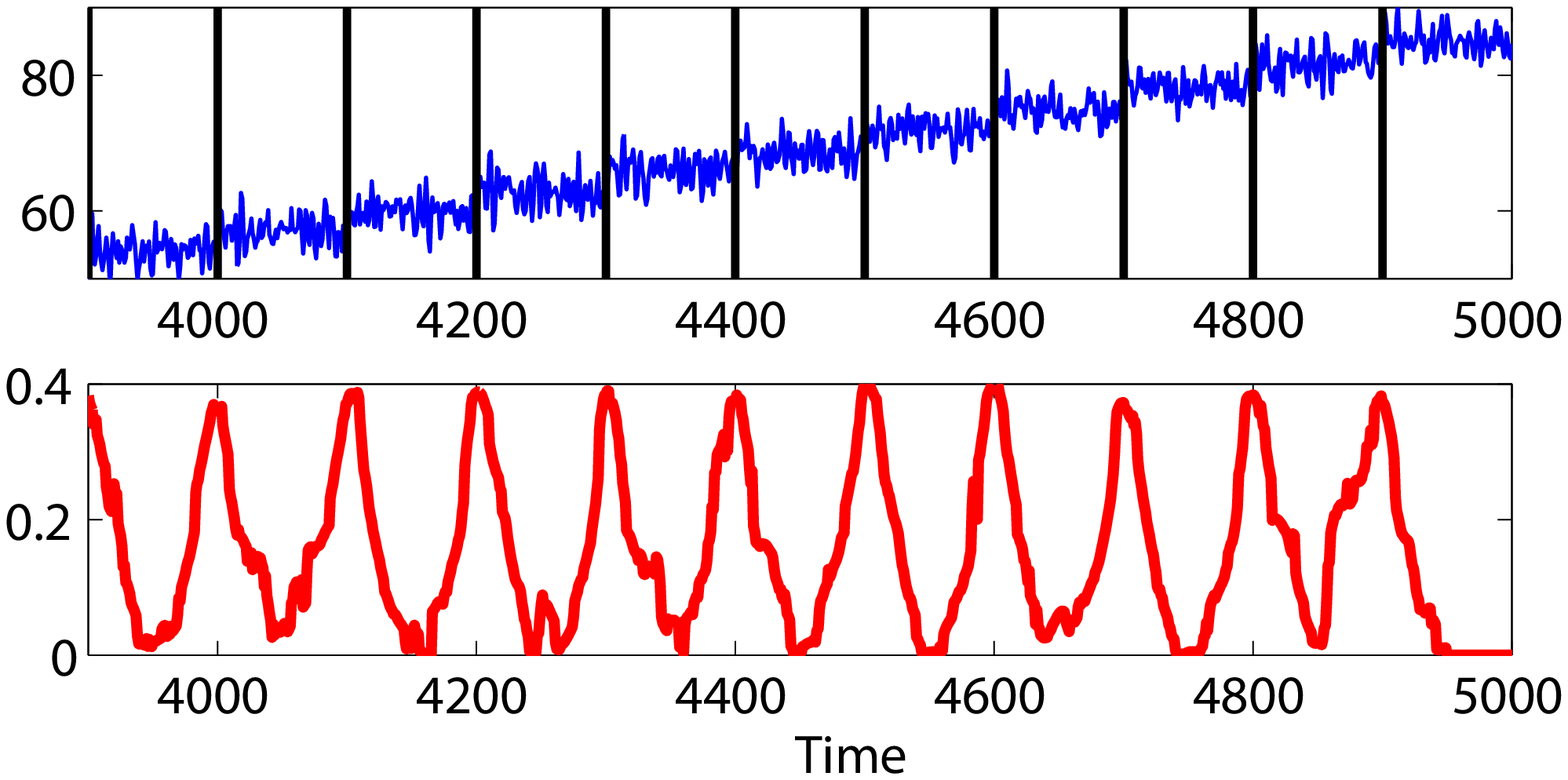}
 \label{fig.illus.dataset1}
 }
 \subfigure[Dataset2]{
 \includegraphics[width=\textwidth,height=.18\textheight,clip]{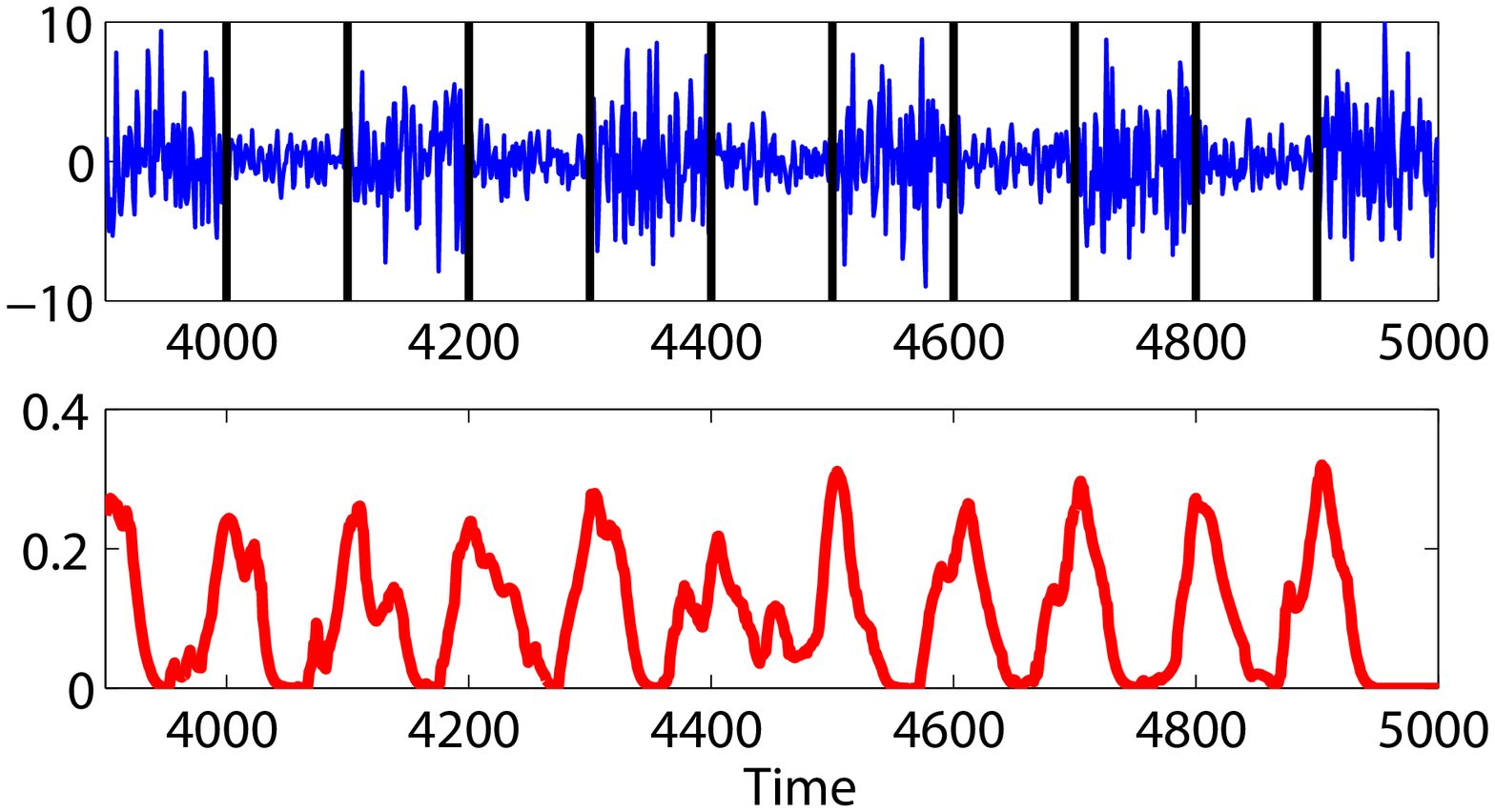}
 \label{fig.illus.dataset2}
 }
 \subfigure[Dataset3]{
 \includegraphics[width=\textwidth,height=.18\textheight,clip]{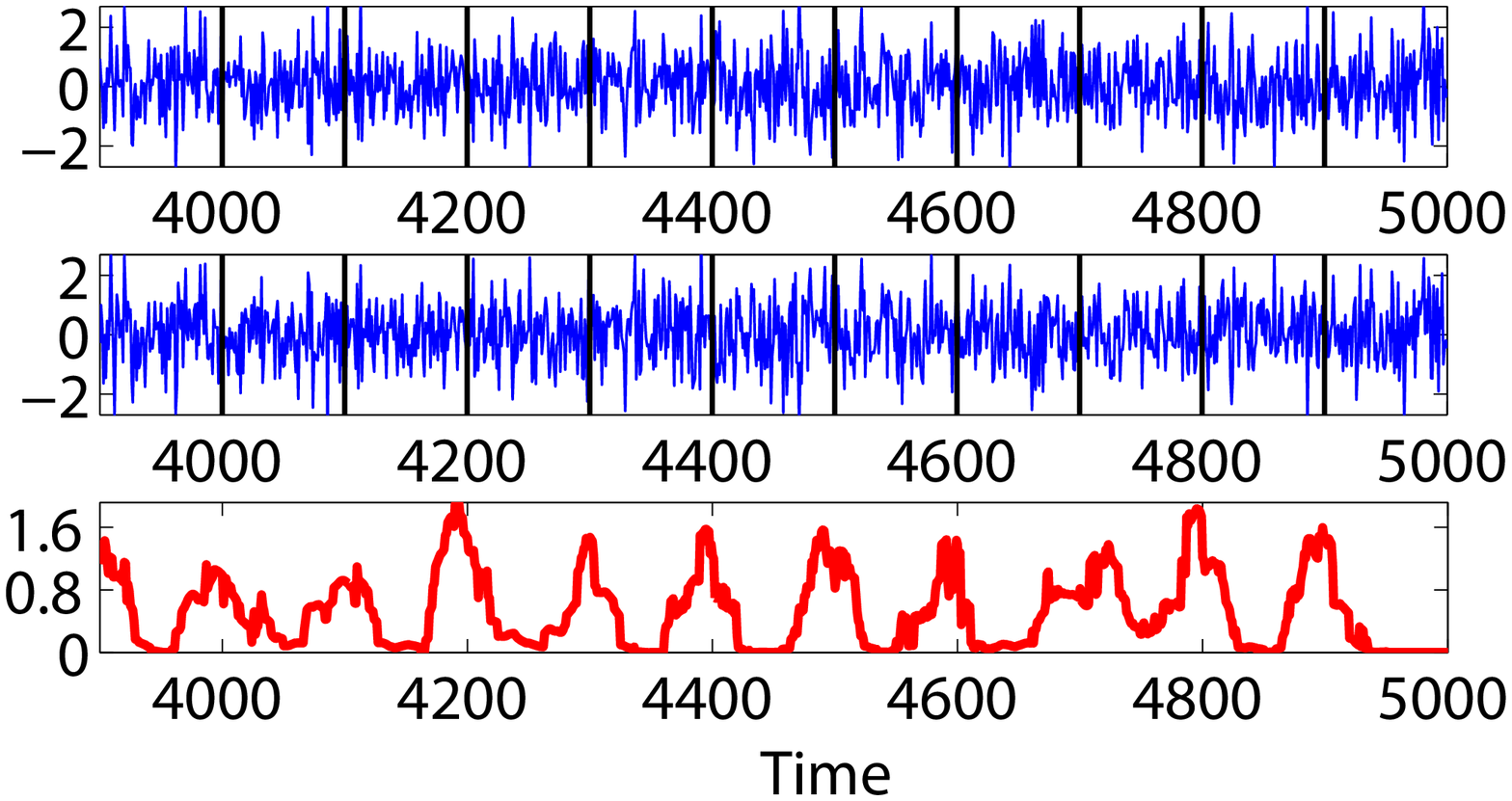}
 \label{fig.illus.dataset3}
 }
 \subfigure[Dataset4]{
 \includegraphics[width=\textwidth,height=.18\textheight,clip]{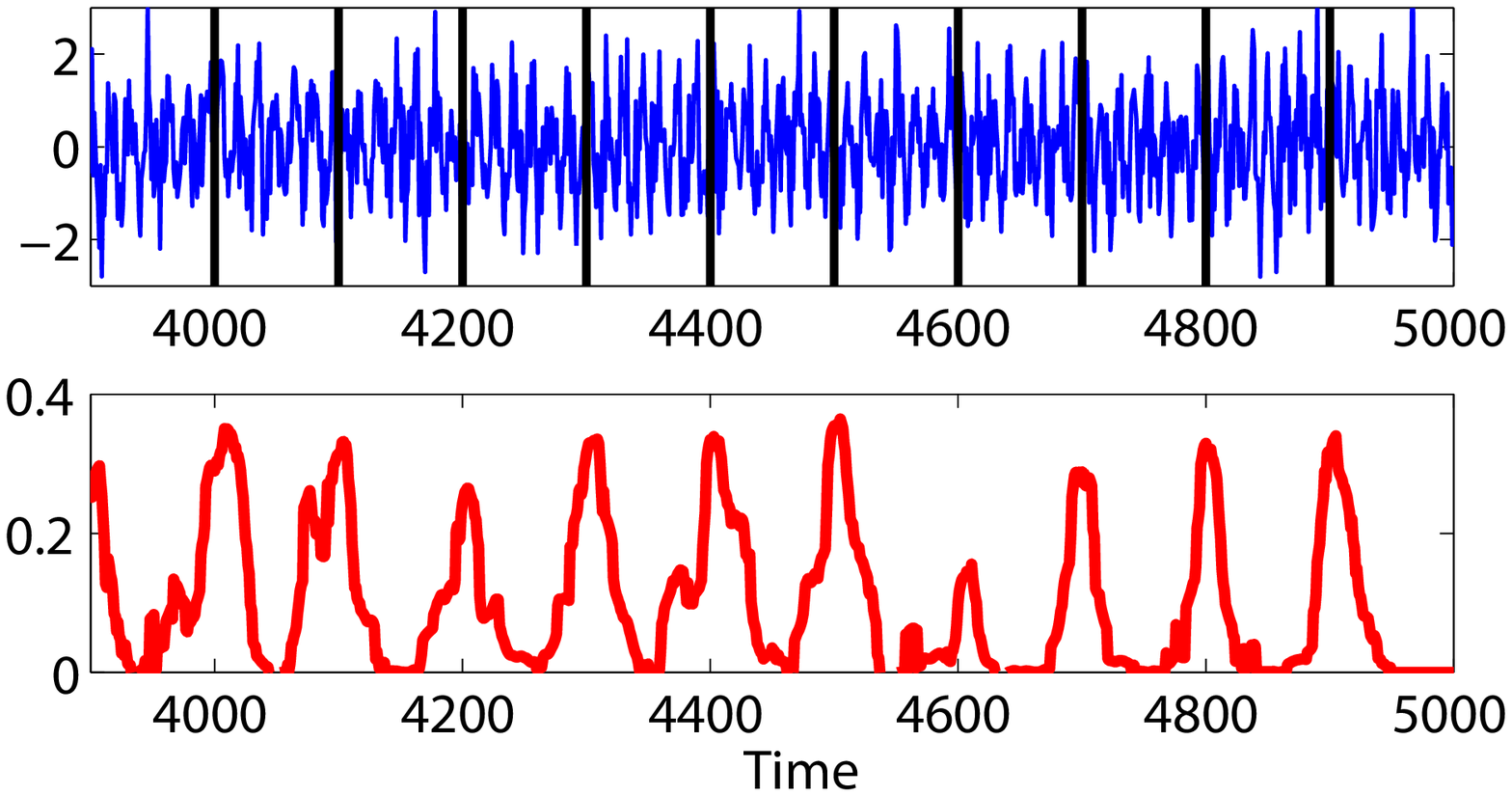}
 \label{fig.illus.dataset4}
 }
\caption{Illustrative time-series samples (upper)
  and the change-point score obtained by the RuLSIF-based method (lower).
  The true change-points are marked by black vertical lines in the upper graphs.}
\label{fig.illus.datasets}
\end{minipage}
~~~~
\begin{minipage}[t]{0.38\textwidth}
\centering
\subfigure[Dataset1]{
 \includegraphics[height=.18\textheight,clip]{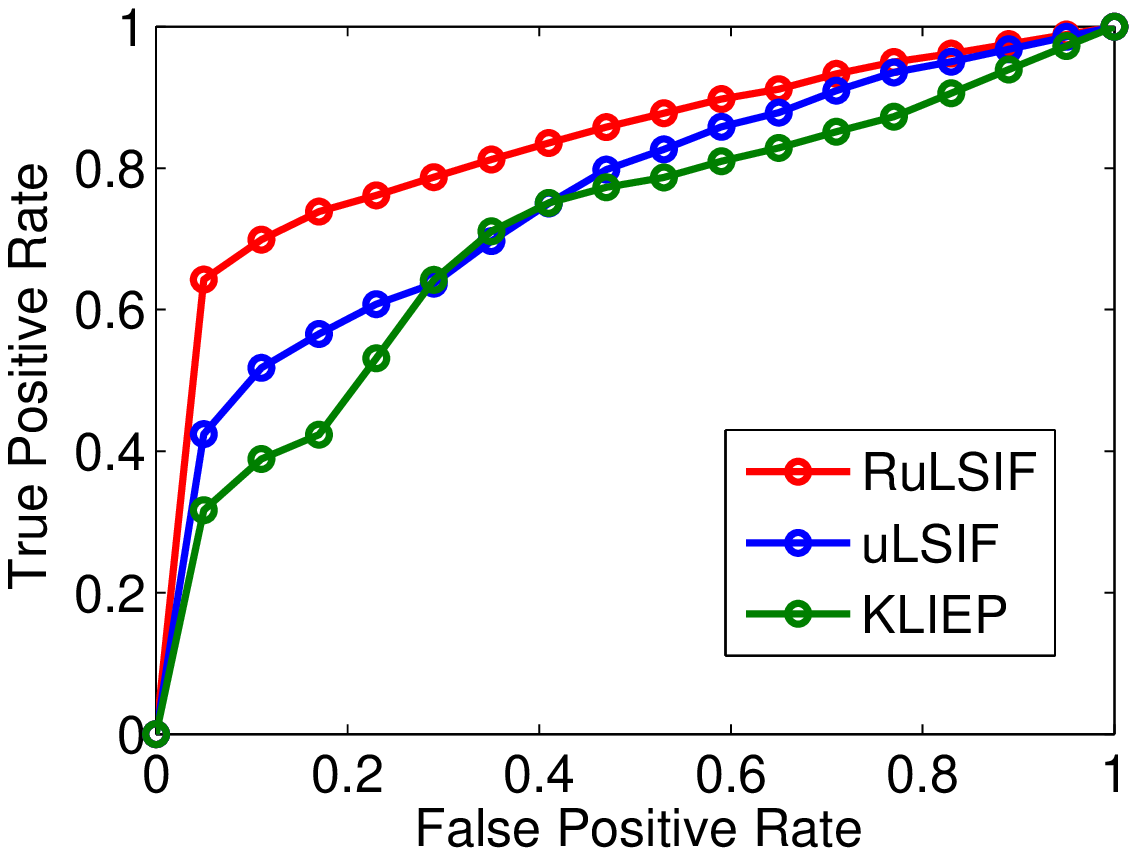}
 \label{fig.roc.dataset1}
 }
 \subfigure[Dataset2]{
 \includegraphics[height=.18\textheight,clip]{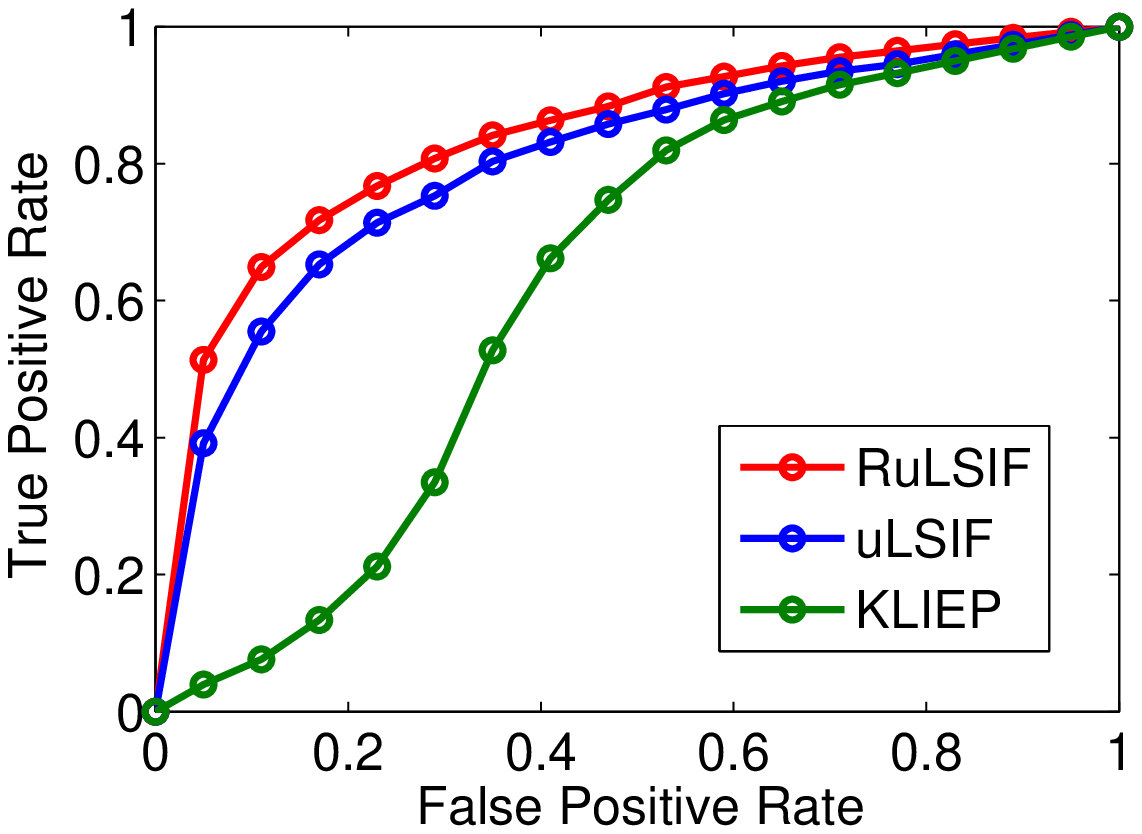}
 \label{fig.roc.dataset2}
 }
 \subfigure[Dataset3]{
 \includegraphics[height=.18\textheight,clip]{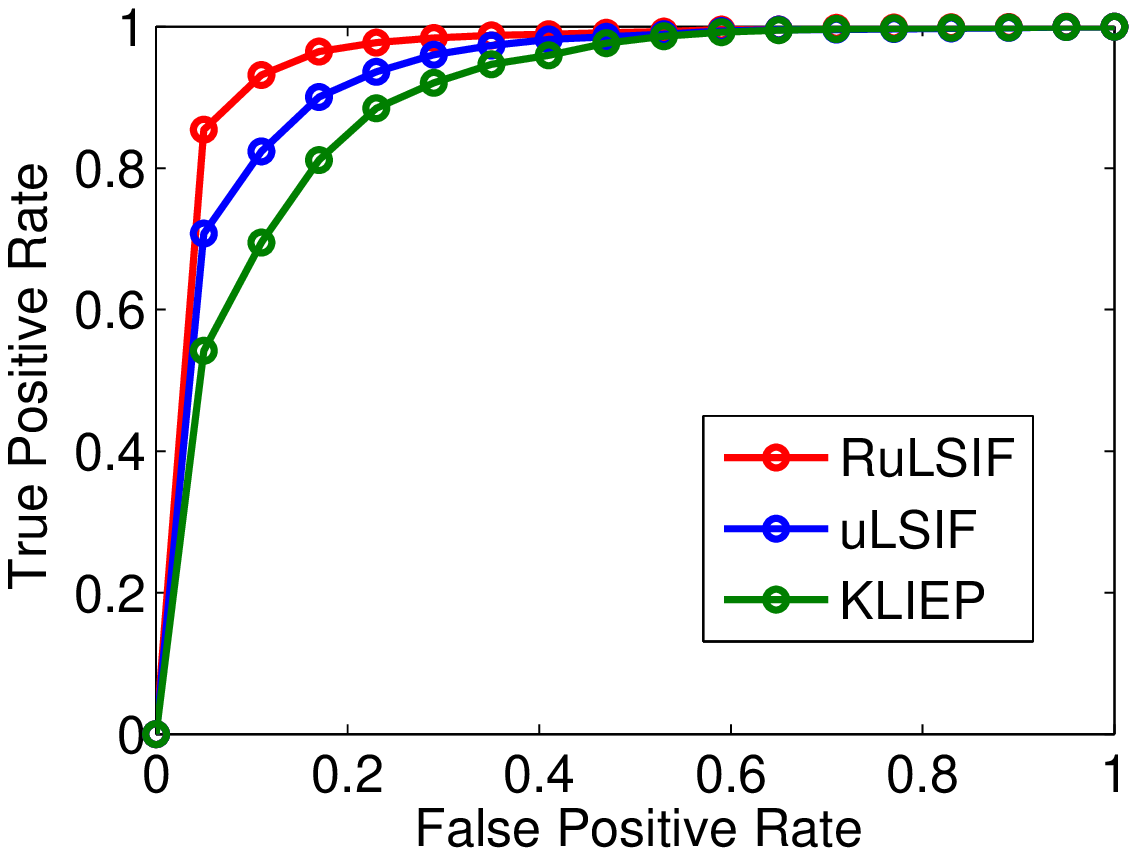}
 \label{fig.roc.dataset3}
 }
 \subfigure[Dataset4]{
 \includegraphics[height=.18\textheight,clip]{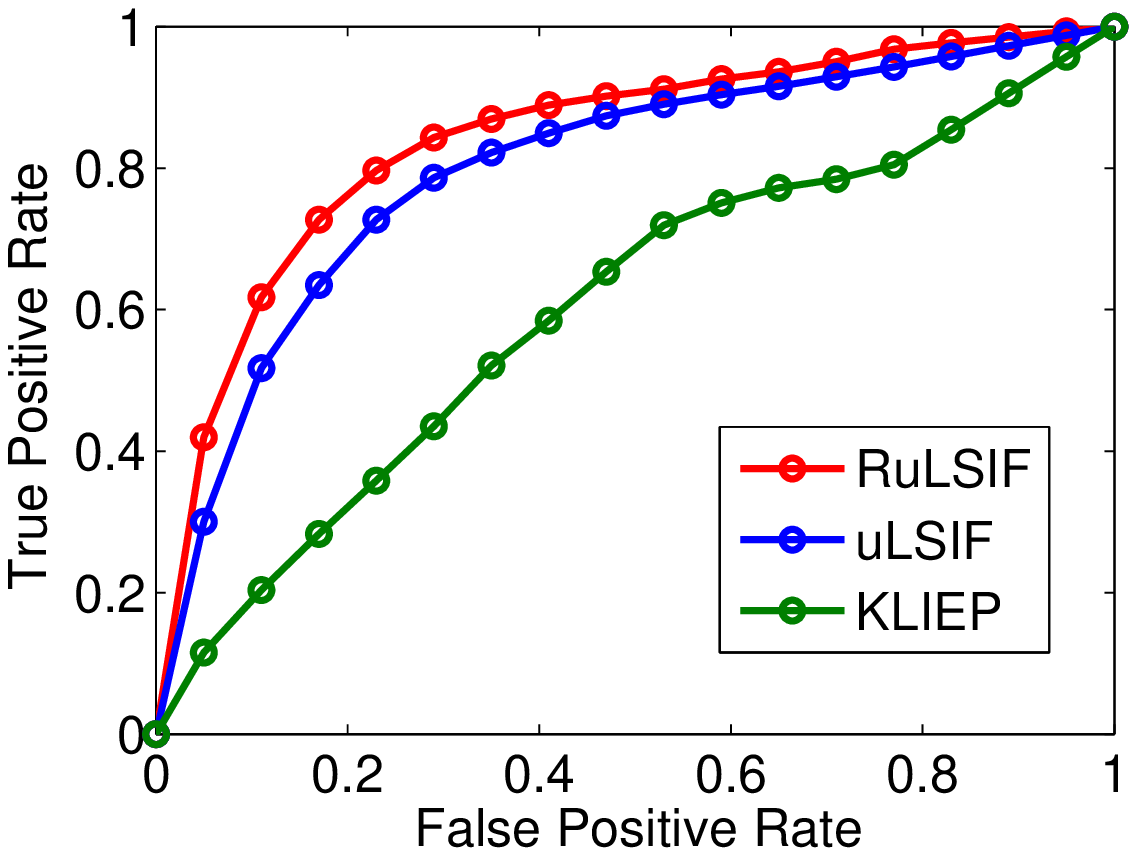}
 \label{fig.roc.dataset4}
 }
\caption{Average ROC curves of RuLSIF-based, uLSIF-based, and KLIEP-based methods.}
\label{fig.roc.datasets}
\end{minipage}
\end{figure*}

Note that, to explore the ability of
detecting change points with different significance,
we purposely made latter change-points more significant than
earlier ones in the above datasets.


Figure~\ref{fig.illus.datasets} shows examples of these datasets
for the last 10 change points and corresponding
change-point score obtained by the proposed RuLSIF-based method.
Although the last 10 change points are the most significant,
we can see from the graphs that, for Dataset 3 and Dataset 4,
these change points can be even hardly identified by human.
Nevertheless,
the change-point score obtained by the proposed RuLSIF-based method
increases rapidly after changes occur.

\begin{table}
\centering
\caption{The AUC values of RuLSIF-based, uLSIF-based, and KLIEP-based methods.
The best and comparable methods by the t-test with significance level $5\%$
are described in boldface.
}
\vspace*{2mm}
\begin{tabular}{ c | c c c c c }
   & RuLSIF & uLSIF & KLIEP \\ \hline
  Dataset 1 & \textbf{.848(.023)} & .763(.023) & .713(.036)\\
  Dataset 2 & \textbf{.846(.031)} & .806(.035) & .623(.040)\\
  Dataset 3 & \textbf{.972(.012)} & .943(.015) & .904(.017)\\
  Dataset 4 & \textbf{.844(.031)} & .801(.024) & .602(.036)\\
\end{tabular}
\label{tab.exp.artifi}
\end{table}

Next, we compare the performance of
RuLSIF-based, uLSIF-based, and KLIEP-based methods
in terms of the \emph{receiver operating characteristic (ROC) curves}
and the area under the ROC curve (AUC) values. We define the \emph{true positive rate} and \emph{false positive rate} in the following way \citep{CPDT_JOUR}:
\begin{itemize}
\item True positive rate (TPR): $n_\mathrm{{cr}}/n_\mathrm{{cp}}$,
\item False positive rate (FPR): $(n_\mathrm{{al}} - n_\mathrm{{cr}}) /n_\mathrm{{al}}$,
\end{itemize}
where $n_\mathrm{{cr}}$ denotes the number of times change points are correctly detected, $n_\mathrm{{cp}}$ denotes the number of all change points, and $n_\mathrm{{al}}$ is the number of all detection alarms.

Following the strategy of the previous researches \citep{ONESVM_PAPER, Bach_CPDT}, peaks of a change-point score are regarded as detection alarms. 
More specifically, a detection alarm at step $t$ is regarded as correct
if there exists a true alarm at step $t^*$ such that $t \in [t^*-10,t^*+10]$.
To avoid duplication, we remove the $k$th alarm at step $t_k$ if $t_k - t_{k-1} < 20$.

We set up a threshold $\eta$ for filtering out all alarms whose change-point scores are lower than or equal to $\eta$. Initially, we set $\eta$ to be equal to the score of the highest peak. Then, by lowering $\eta$ gradually, both TPR and FPR become non-decreasing. For each $\eta$, we plot TPR and FPR on the graph, and thus a monotone curve can be drawn. 
 

Figure~\ref{fig.roc.datasets} illustrates ROC curves averaged over $50$ runs
with different random seeds for each dataset.
Table~\ref{tab.exp.artifi} describes the mean and standard deviation of the AUC values
over $50$ runs.
The best and comparable methods by the t-test with significance level $5\%$
are described in boldface.
The experimental results show that the uLSIF-based method tends to outperform
the KLIEP-based method,
and the RuLSIF-based method even performs better than the uLSIF-based method. 

\begin{figure}
\subfigure[Dataset 1 ($n = 25$)]{
 \includegraphics[width=.31\textwidth,clip]{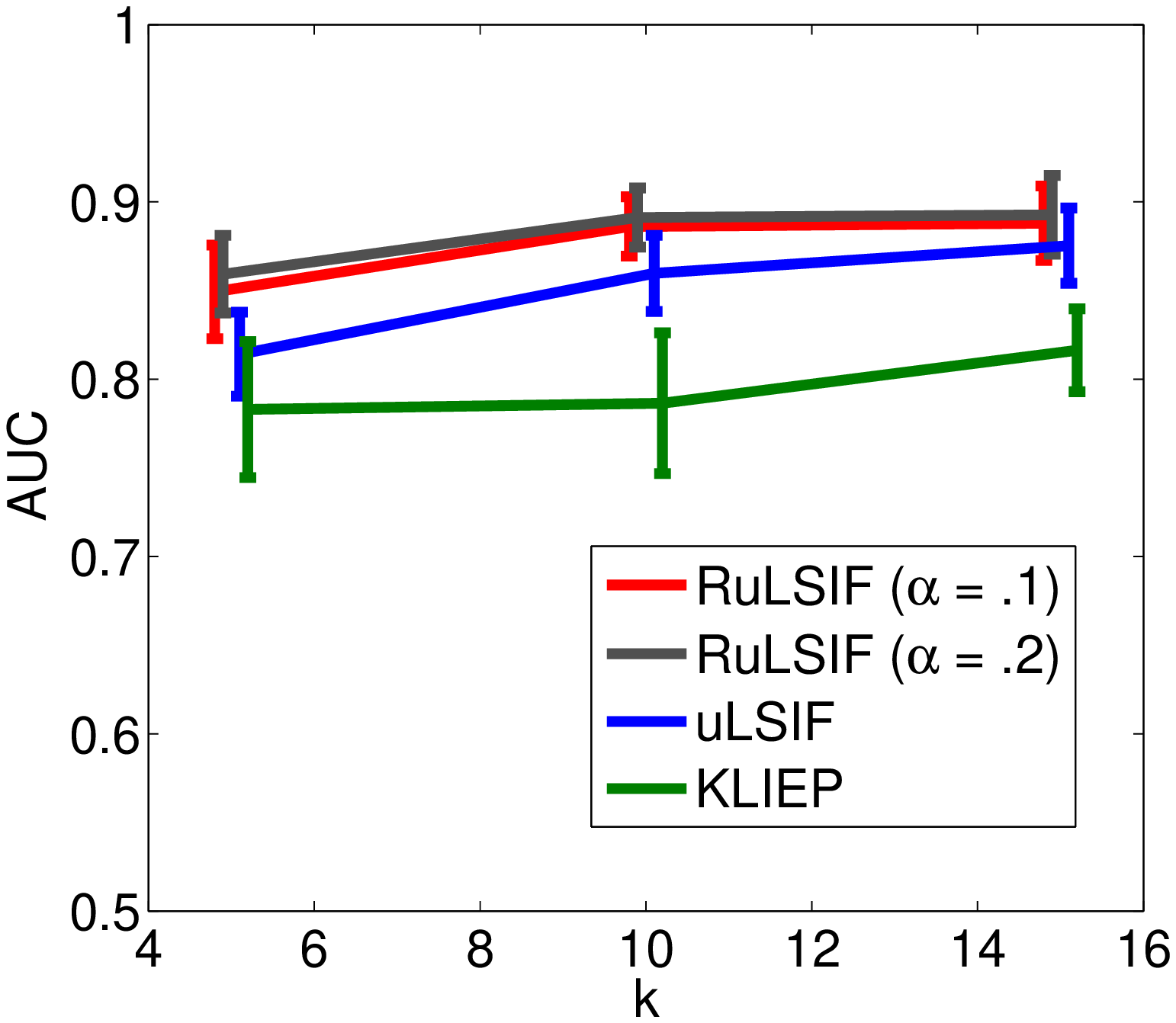}
 }
\subfigure[Dataset 1 ($n = 50$)]{
 \includegraphics[width=.31\textwidth,clip]{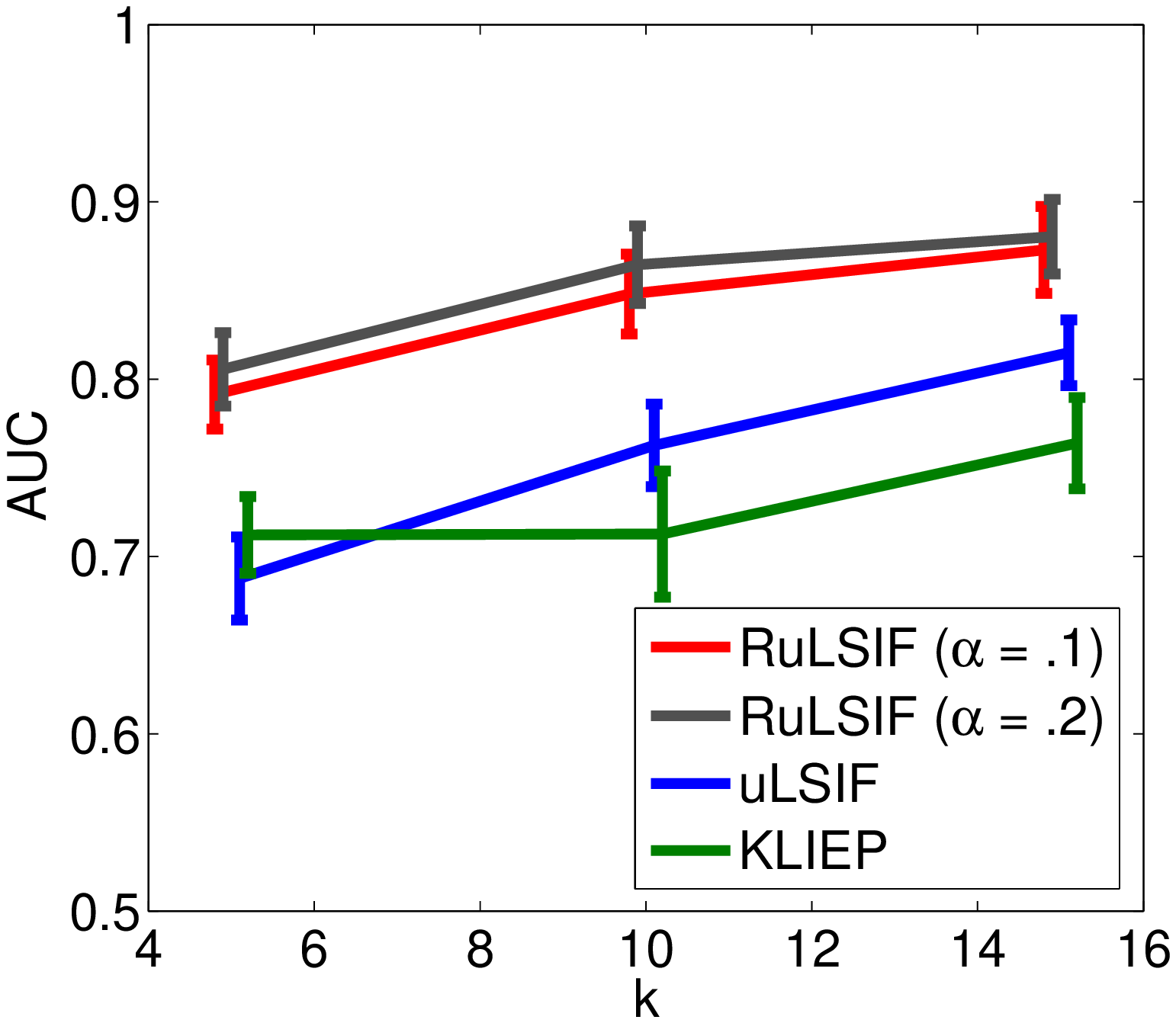}
 }
\subfigure[Dataset 1 ($n = 75$)]{
 \includegraphics[width=.31\textwidth,clip]{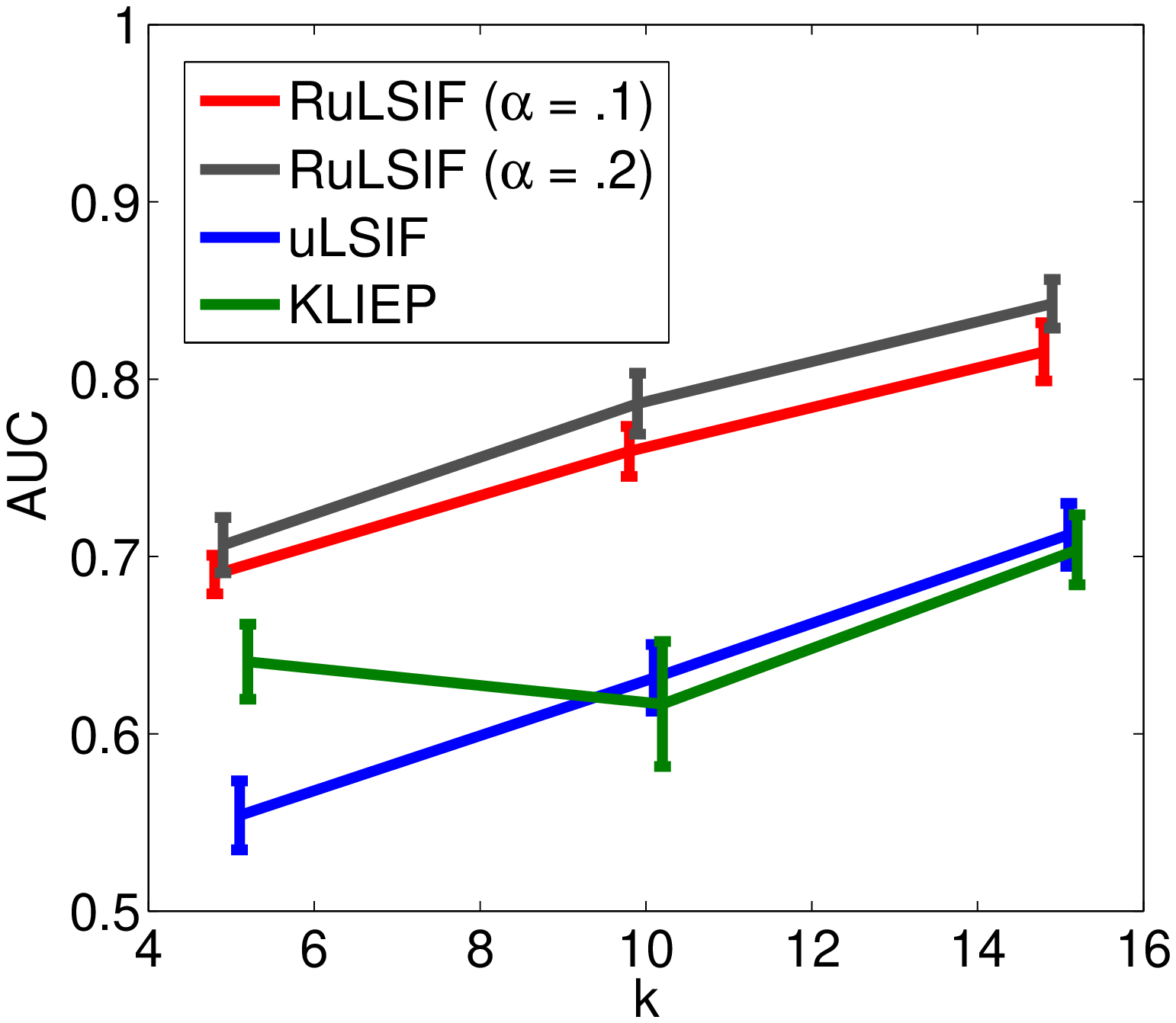}
 }\\
 \subfigure[Dataset 2 ($n = 25$)]{
 \includegraphics[width=.31\textwidth,clip]{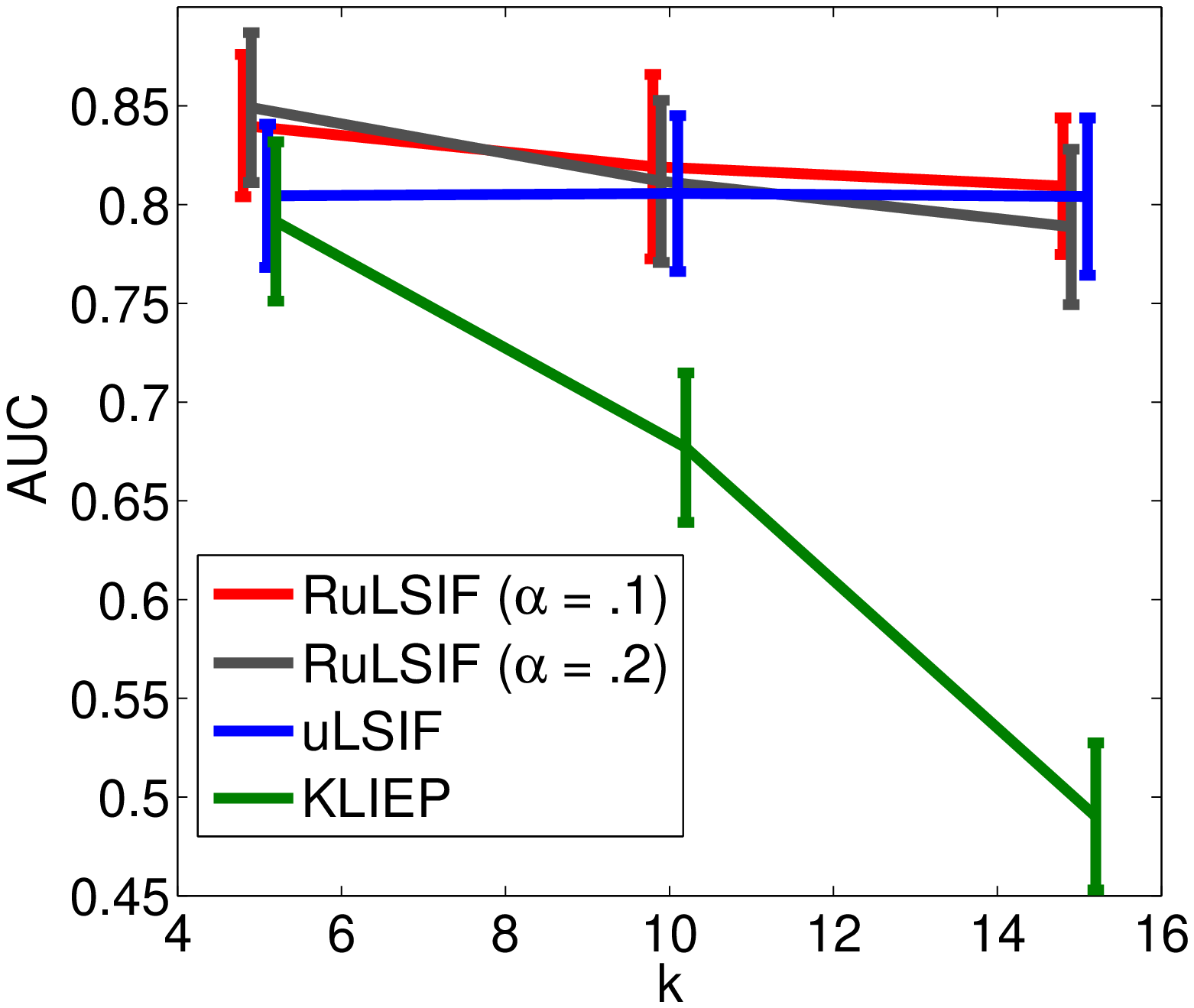}
 }
\subfigure[Dataset 2 ($n = 50$)]{
 \includegraphics[width=.31\textwidth,clip]{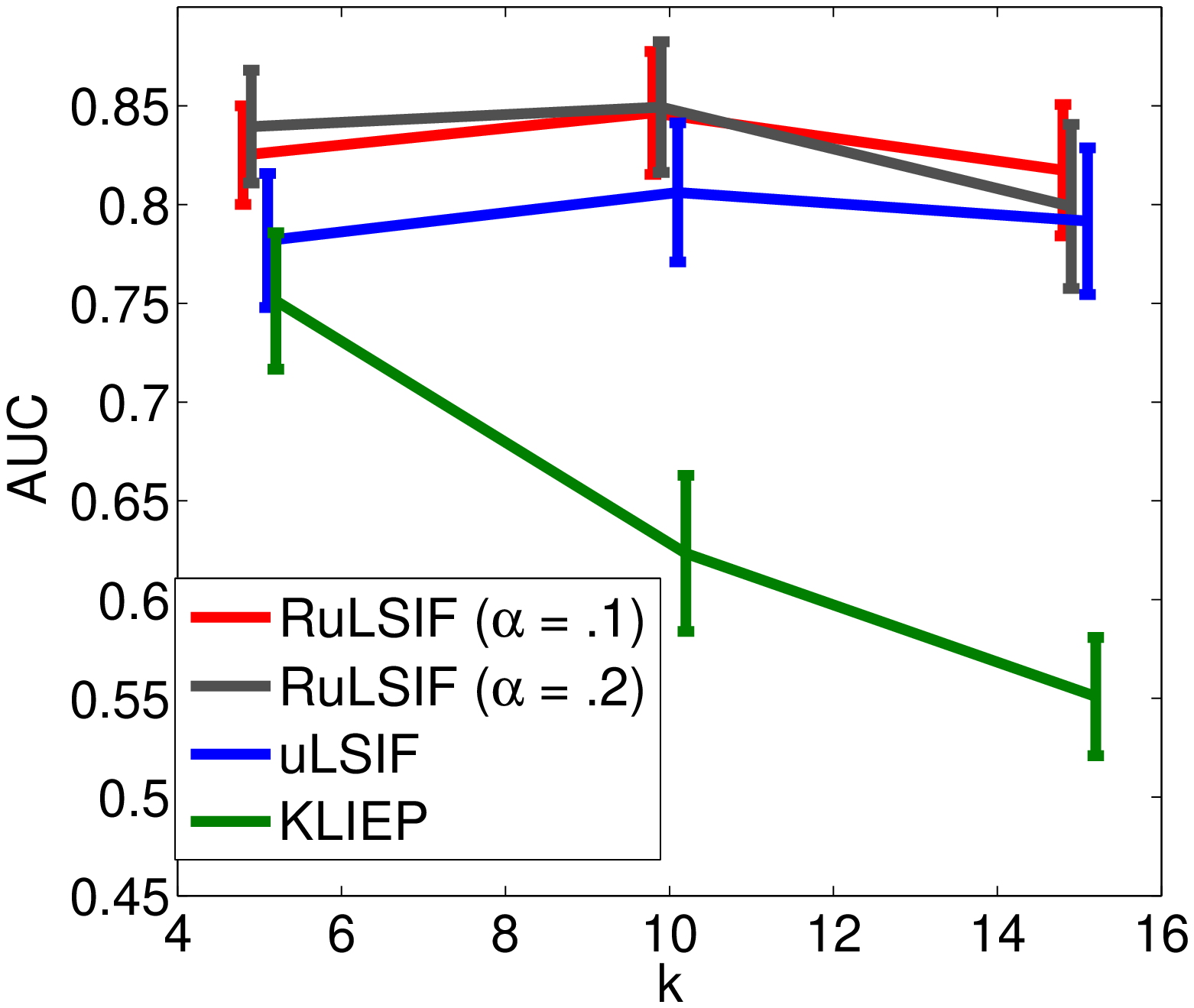}
 }
\subfigure[Dataset 2 ($n = 75$)]{
 \includegraphics[width=.31\textwidth,clip]{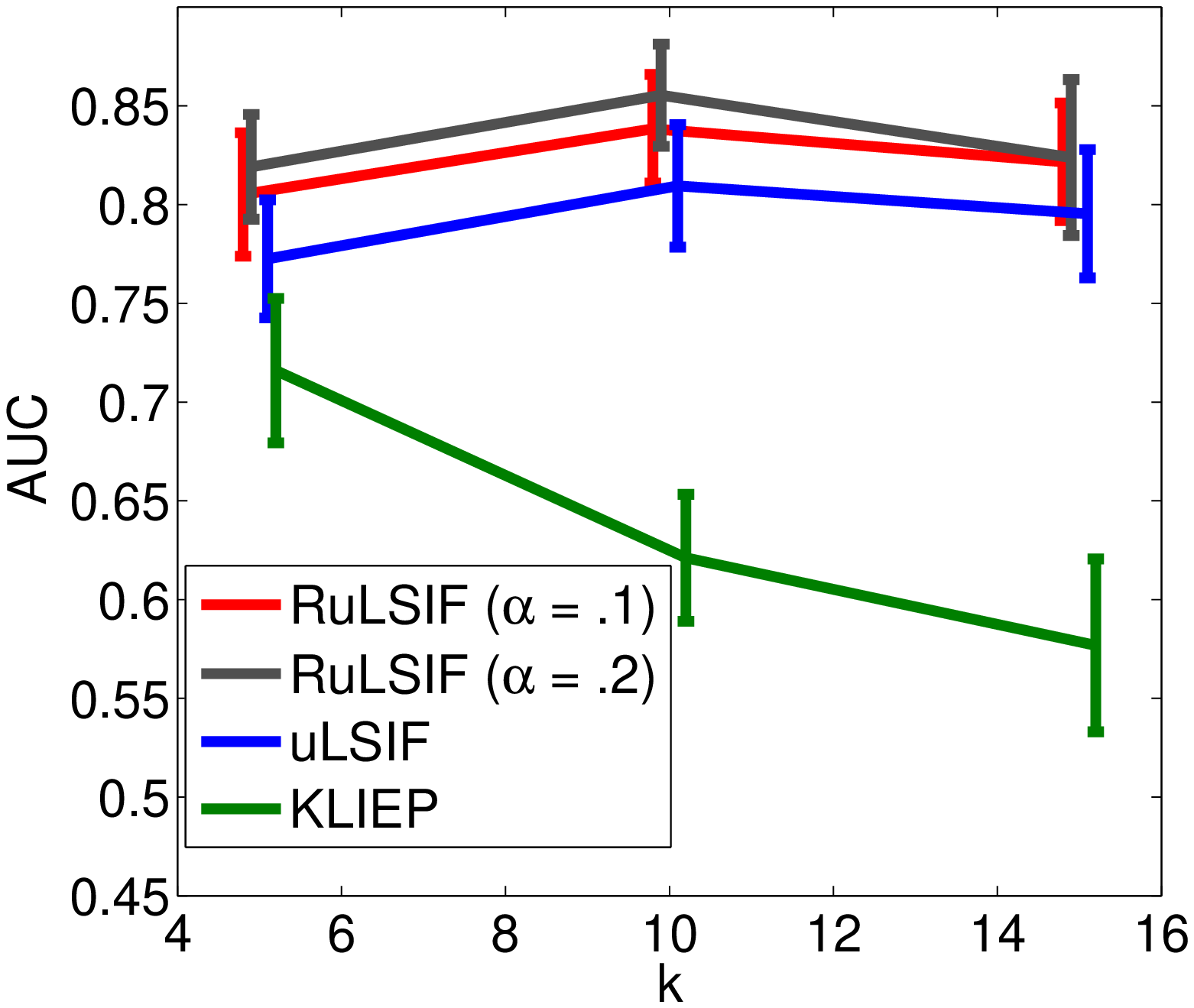}
 }\\
 \subfigure[Dataset 3 ($n = 25$)]{
 \includegraphics[width=.31\textwidth,clip]{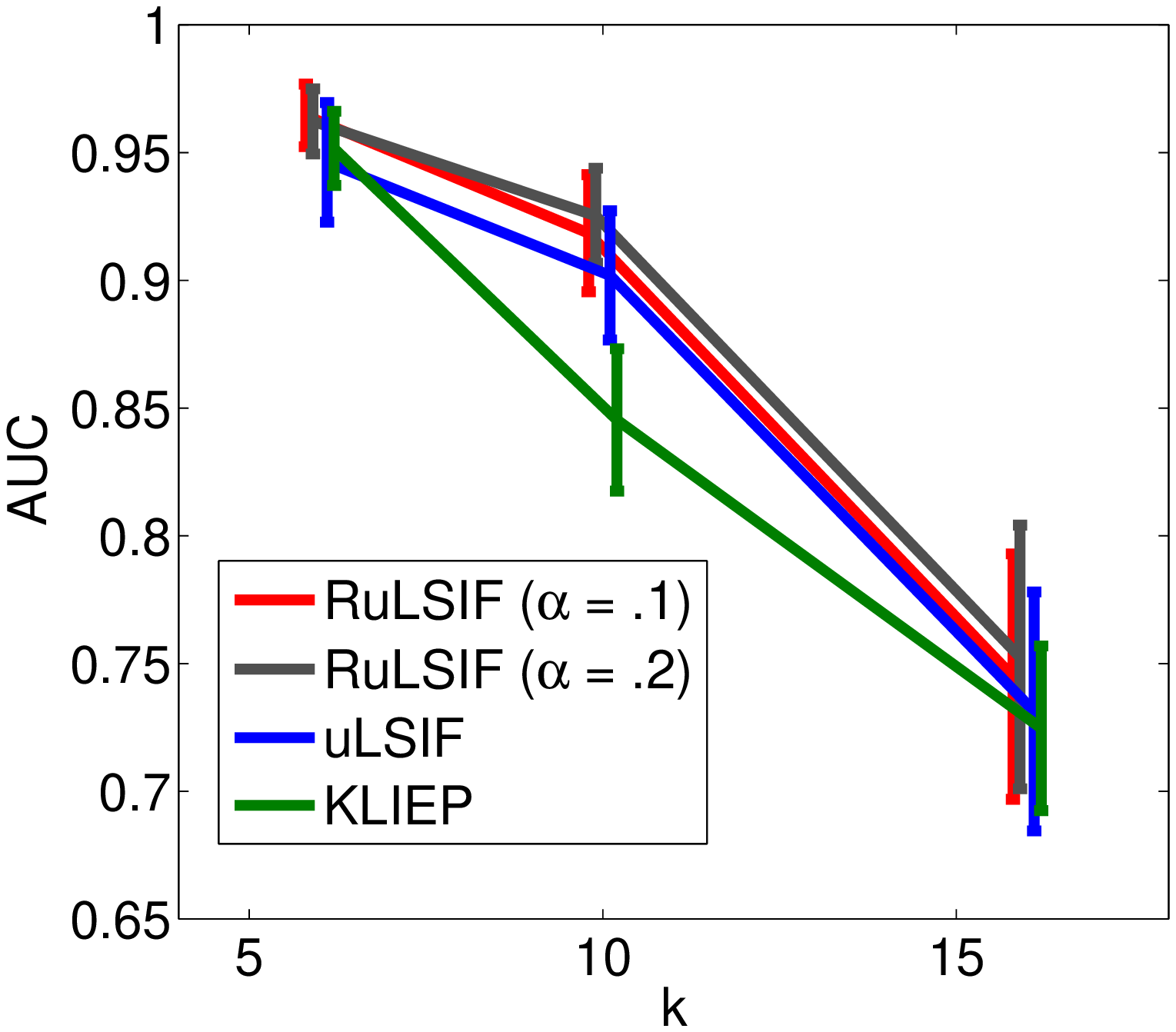}
 }
\subfigure[Dataset 3 ($n = 50$)]{
 \includegraphics[width=.31\textwidth,clip]{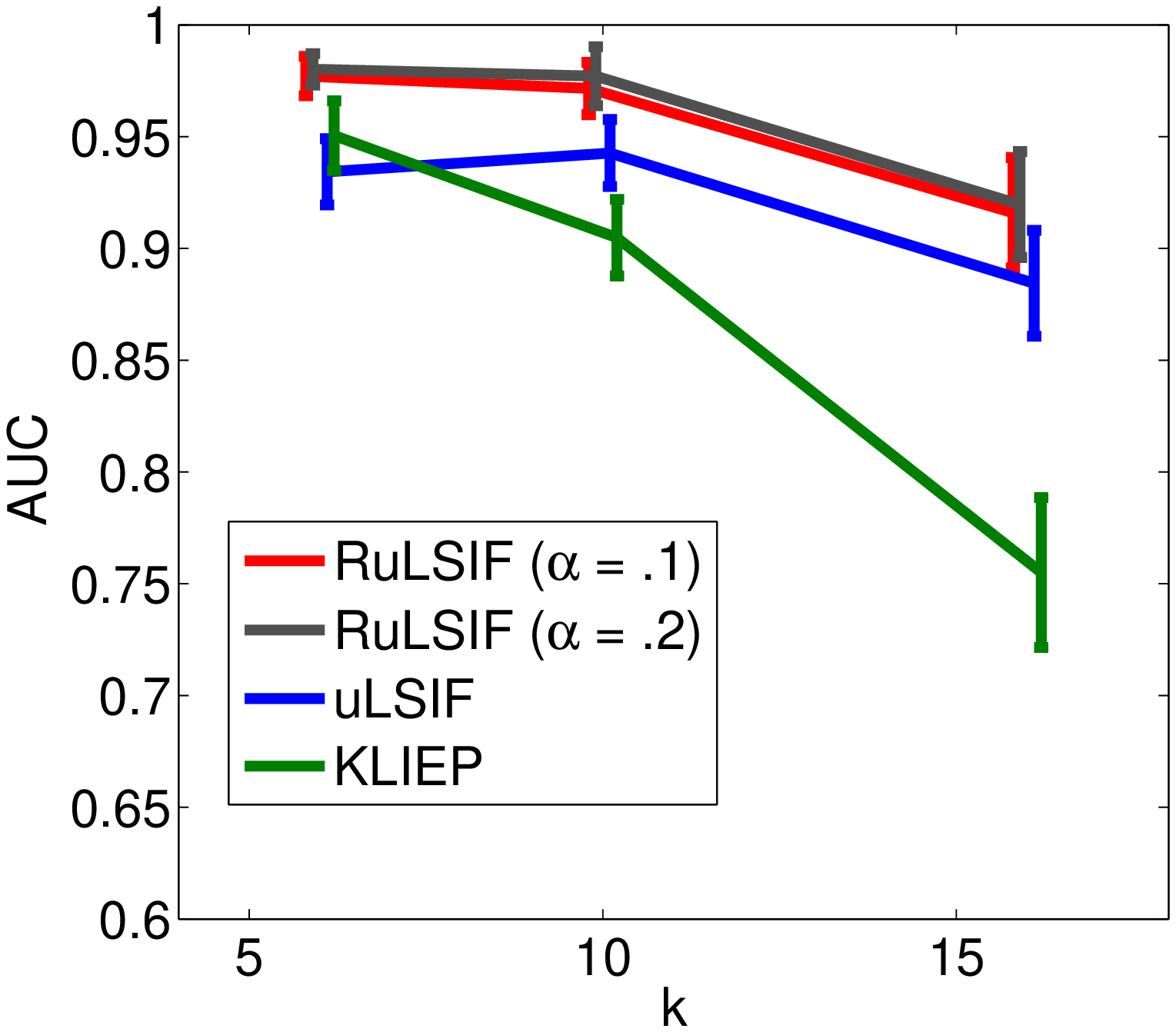}
 }
\subfigure[Dataset 3 ($n = 75$)]{
 \includegraphics[width=.31\textwidth,clip]{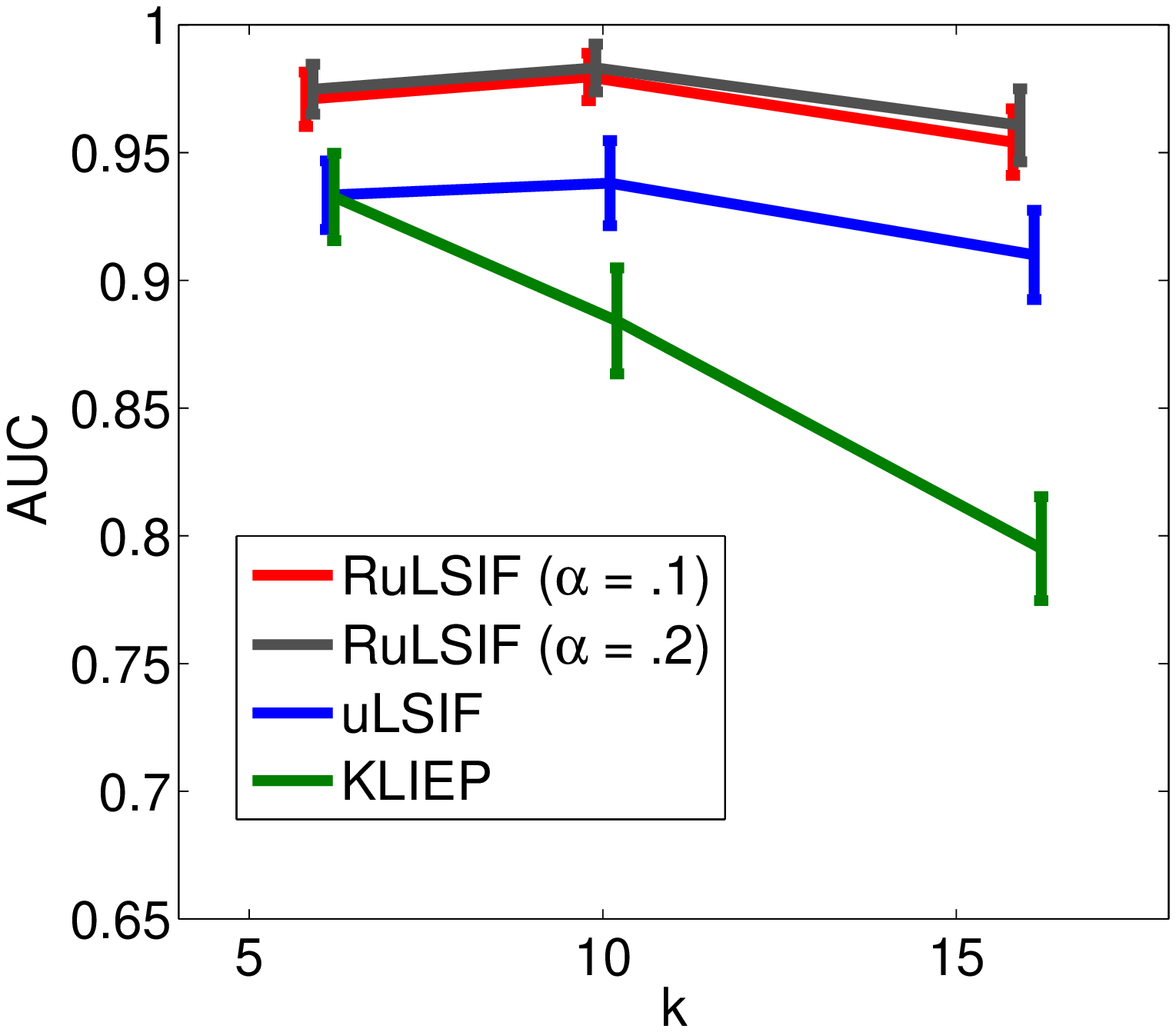}
 }\\
 \subfigure[Dataset 4 ($n = 25$)]{
 \includegraphics[width=.31\textwidth,clip]{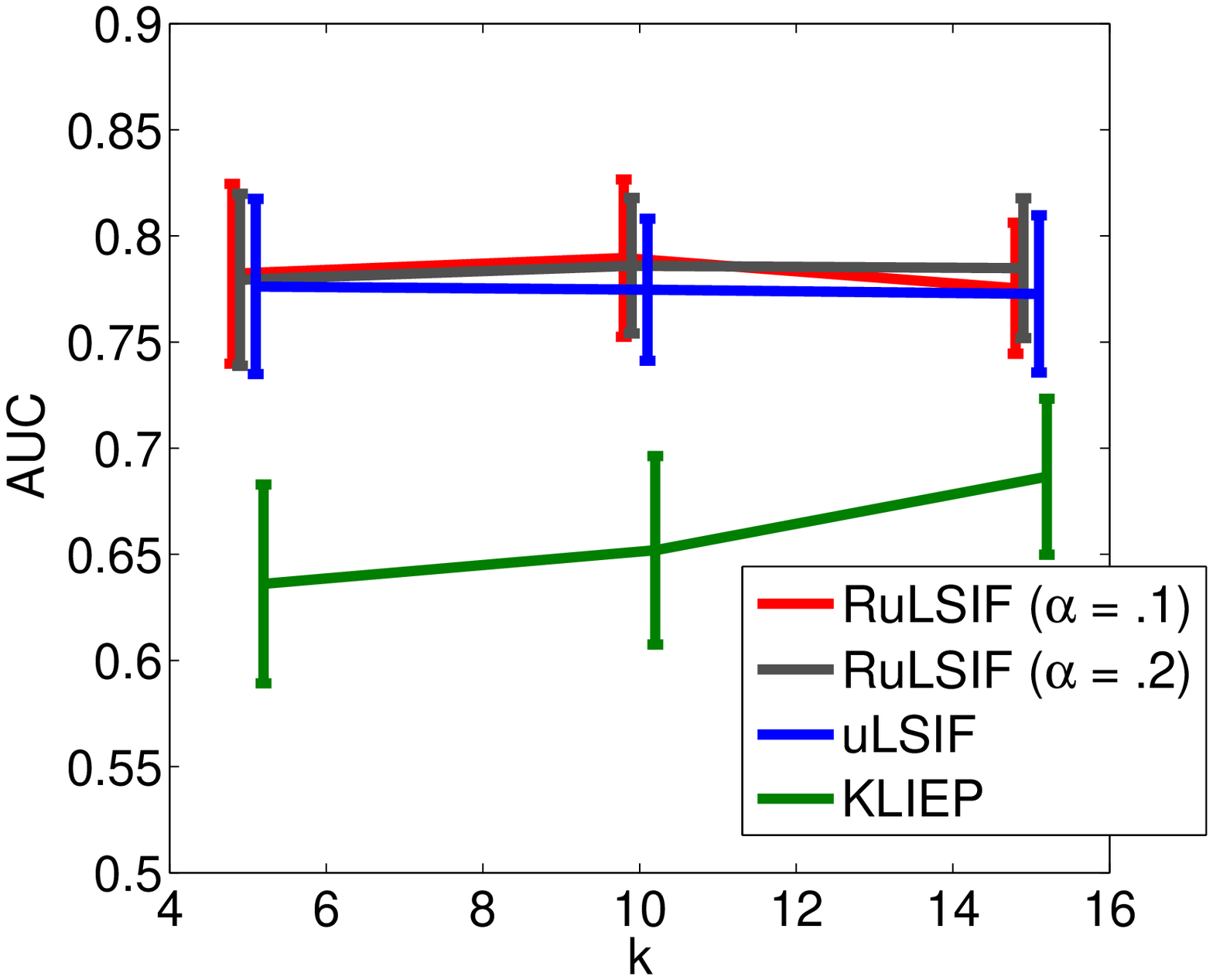}
 }
\subfigure[Dataset 4 ($n = 50$)]{
 \includegraphics[width=.31\textwidth,clip]{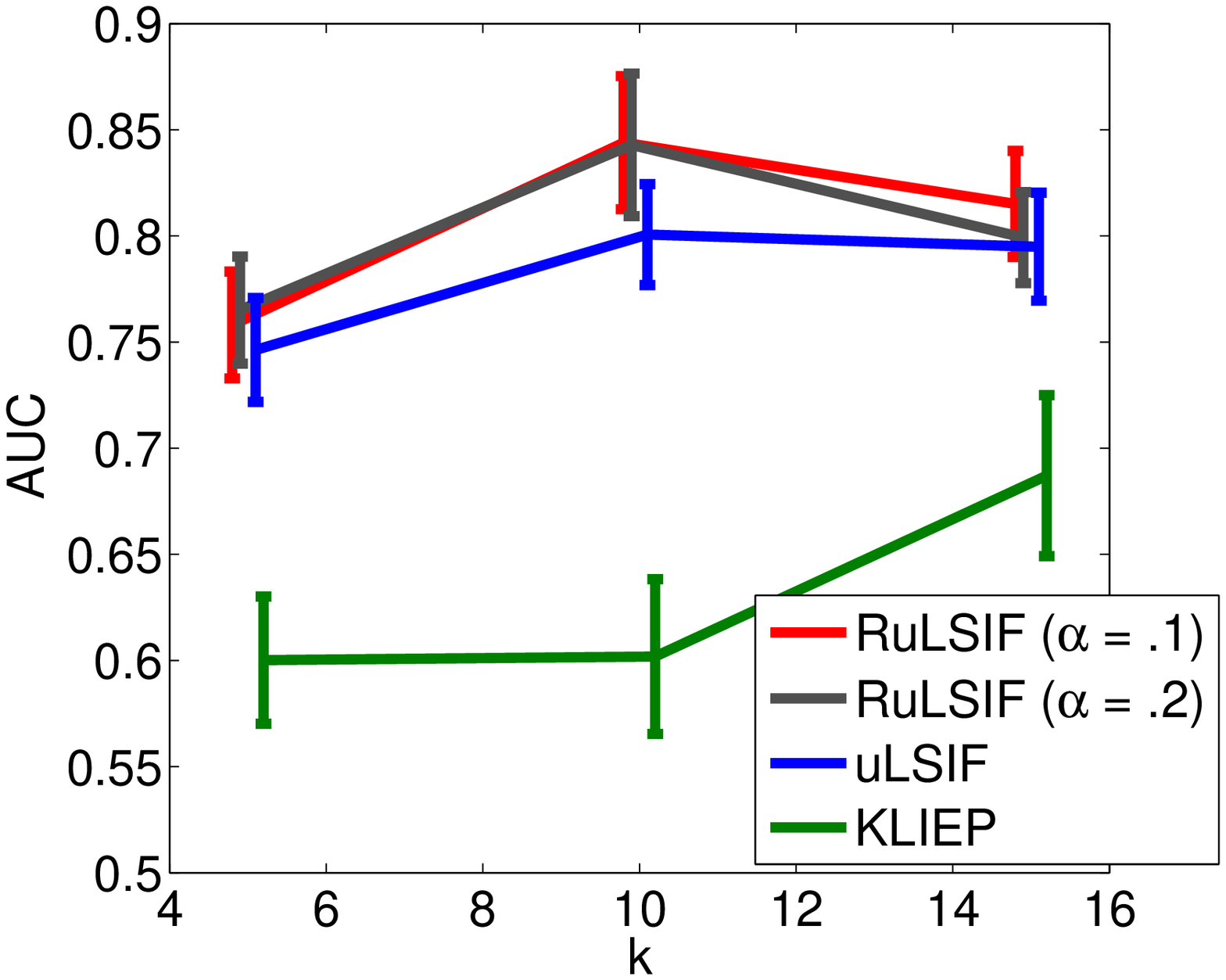}
 }
\subfigure[Dataset 4 ($n = 75$)]{
 \includegraphics[width=.31\textwidth,clip]{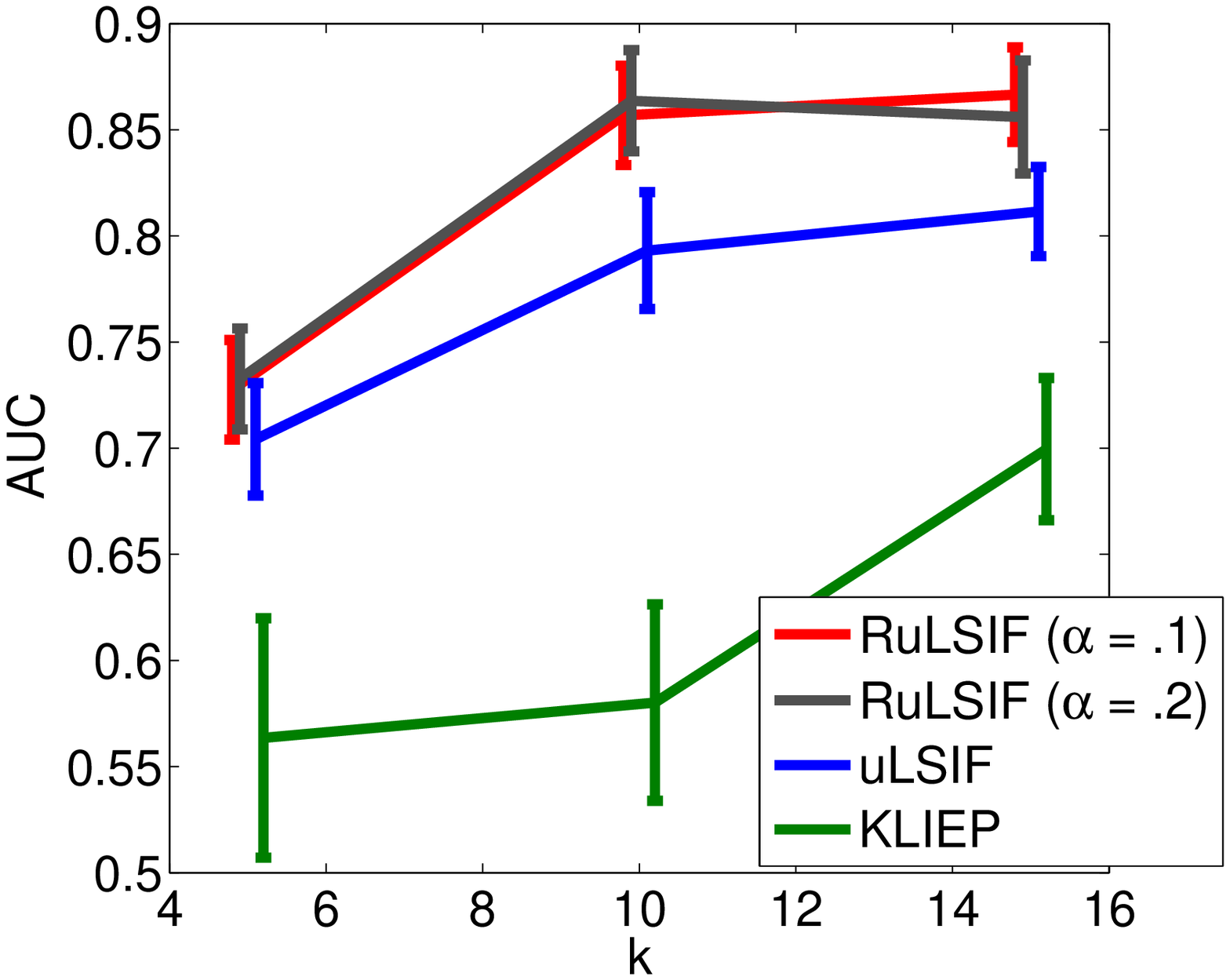}
 }\\
 \caption{AUC plots for $ n = 25, 50, 75$ and $k = 5, 10, 15$.
 The horizontal axes denote $k$, while the vertical axes denote AUC values.}
 \label{fig.illus.n.k.AUC}
\end{figure}

Finally, we investigate the sensitivity of the performance on different choices of $n$ and $k$ in terms of AUC values. In Figure~\ref{fig.illus.n.k.AUC}, the AUC values of RuLSIF ($\alpha = 0.1$ and $0.2$), uLSIF (which corresponds to RuLSIF with $\alpha=0$), and KLIEP were plotted for $k = 5$, $10$, and $15$ under a specific choice of $n$ in each graph. We generate such graphs for all 4 datasets with $n = 25$, $50$, and $75$. The result shows that the proposed method consistently performs better than the other methods, and the order of the methods according to the performance is kept unchanged over various choices of $n$ and $k$. Moreover, the RuLSIF methods with $\alpha=0.1$ and $0.2$ perform rather similarly.
For this reason, we keep using the medium parameter values among the candidates in the following experiments:
$n = 50$, $k =  10$, and $\alpha=0.1$.

\subsection{Real-World Datasets}

Next, we evaluate the performance of the density-ratio estimation based methods and
other existing change-point detection methods using two real-world datasets:
Human-activity sensing and speech.

We include the following methods in our comparison.
\begin{itemize}
\item \textbf{Singular spectrum transformation (SST)
    \citep{SST_PAPER,SST2_PAPER,SST_CLIMATE_PAPER}:}
  Change-point scores are evaluated on two consecutive trajectory matrices
  using the distance-based singular spectrum analysis.
  This corresponds to a state-space model with no system noise. For this method, we use the first $4$ eigenvectors to compare the
difference between two subspaces, which was confirmed to be reasonable choice in our preliminary experiments.

\item \textbf{Subspace identification (SI) \citep{SI_PAPER}:}
  SI identifies a subspace in which time-series data is constrained,
  and evaluates the distance of target sequences from the subspace.
  The subspace spanned by the columns of an observability matrix is used
  for estimating the distance from the subspace spanned by subsequences of time-series data. For this method, we use the top $4$ significant singular values according to our preliminary experiment results.

\item \textbf{Auto regressive (AR) \citep{AR_JOUR}:}
  AR first fits an AR model to time-series data, and then auxiliary time-series is
  generated from the AR model. With an extra AR model-fitting,
  the change-point score is given by the log-likelihood. The order of the AR model is chosen by Schwarz's Bayesian information criterion \citep{SBC}.

\item \textbf{One-class support vector machine (OSVM) \citep{ONESVM_PAPER}:}
  Change-point scores are calculated by OSVM using two sets of descriptors of signals. The kernel width $\sigma$ is set to the median value of the distances between samples, which is a popular heuristic in kernel methods \citep{Scholkopf02}. Another parameter $\nu$ is set to 0.2, which indicates the proportion of outliers.
\end{itemize}

First, we use a human activity dataset.
This is a subset of the \emph{Human Activity Sensing Consortium (HASC) challenge
2011}\footnote{\url{http://hasc.jp/hc2011/}},
which provides human activity information collected by portable three-axis accelerometers.
The task of change-point detection is to segment the time-series data
according to the 6 behaviors: ``\textit{stay}'', ``\textit{walk}'', ``\textit{jog}'',
``\textit{skip}'', ``\textit{stair up}'', and ``\textit{stair down}''.
The starting time of each behavior is arbitrarily decided by each user.
Because the orientation of accelerometers is not necessarily fixed,
we take the $\ell_2$-norm of the 3-dimensional (i.e., $x$-, $y$-, and $z$-axes) data.

\begin{figure*}[p]

\begin{minipage}[t]{\textwidth}
\centering
\subfigure[One of the original signals and change-point scores obtained by the RuLSIF-based method]{
 \includegraphics[width=.8\textwidth,clip]{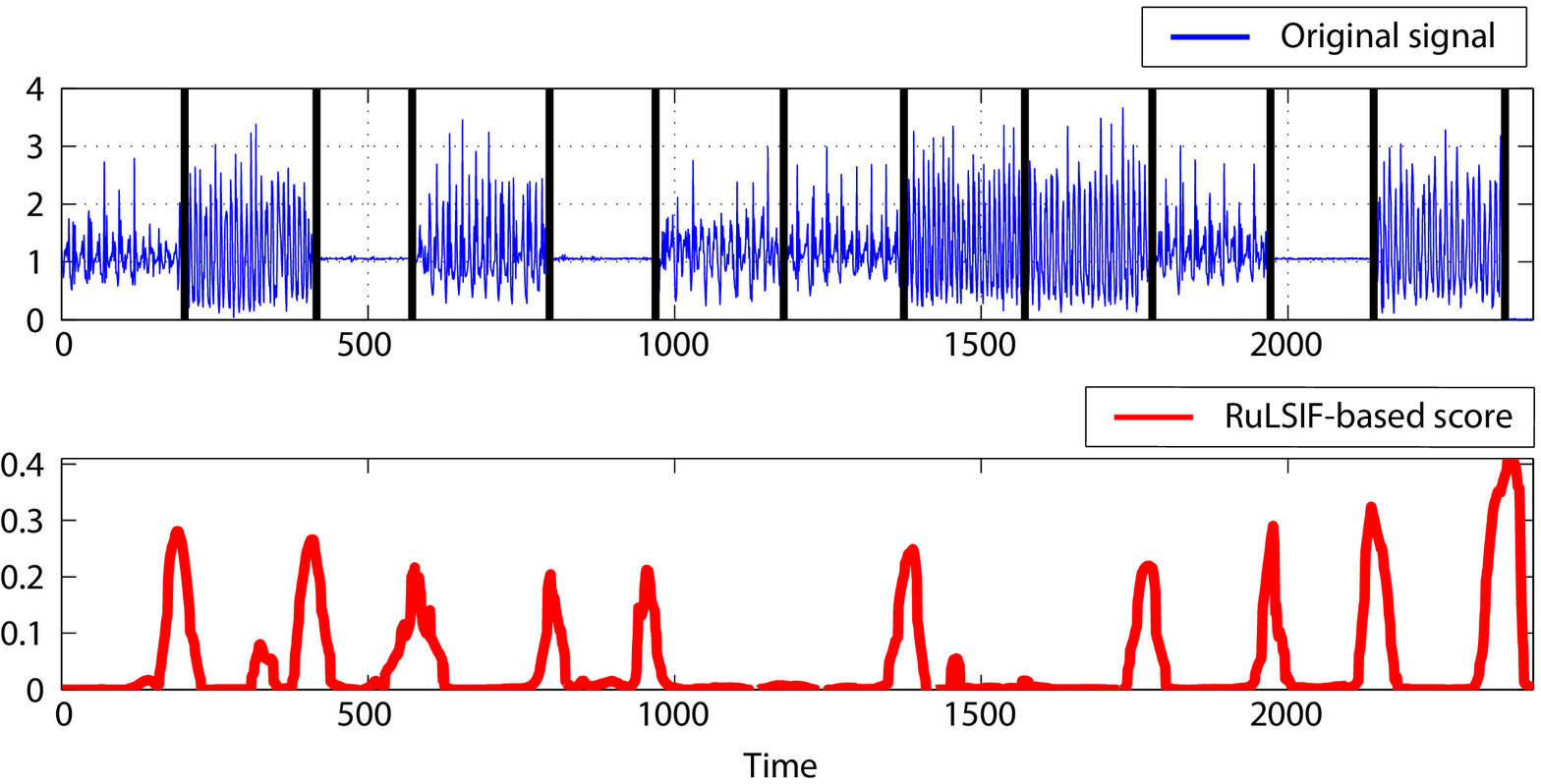}
 \label{fig.illus.human}
 }
 \subfigure[Average ROC curves]{
 \includegraphics[width=.5\textwidth,clip]{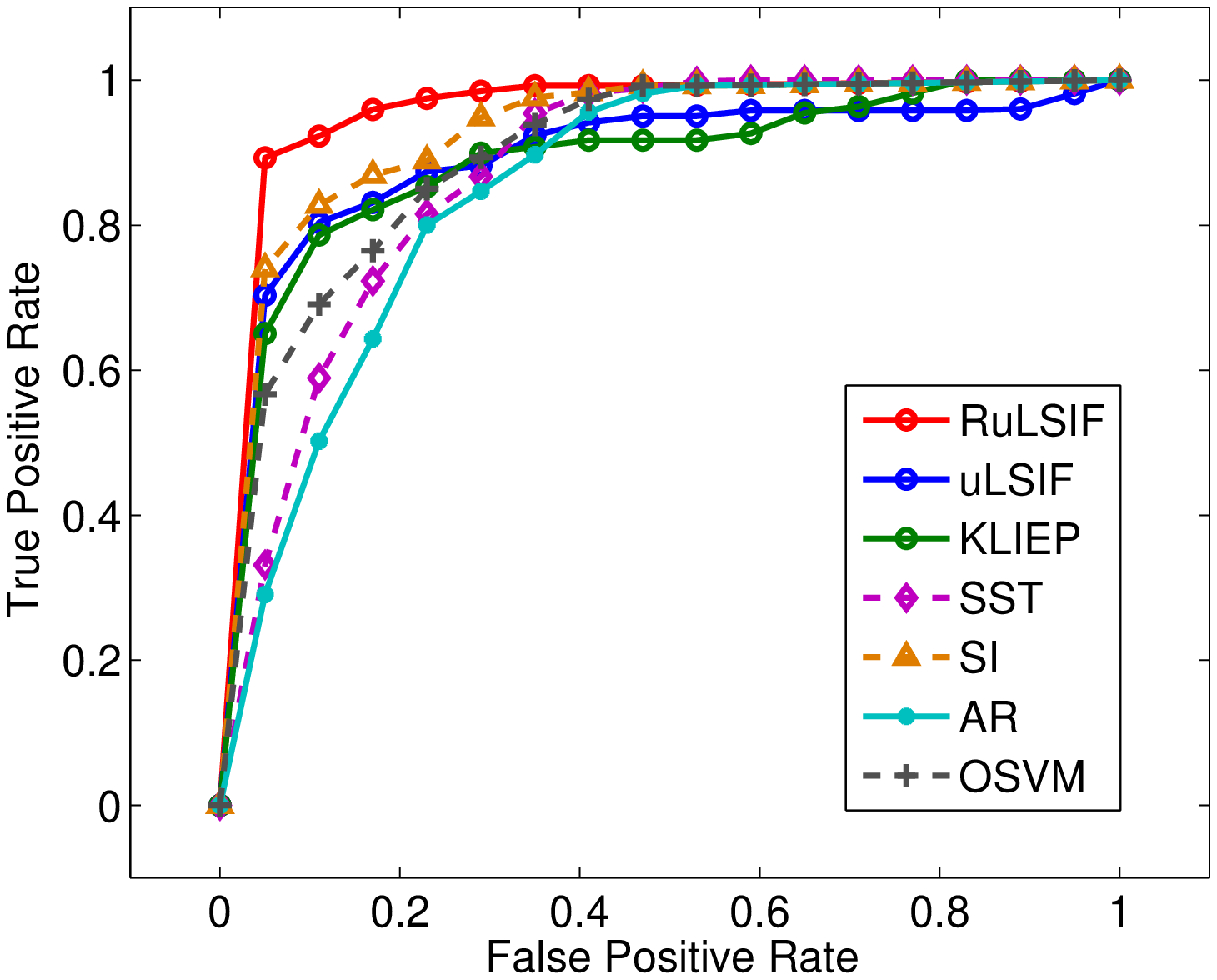}
 \label{fig.roc.human}
 }\\
 \centering
 \subfigure[AUC values. The best and comparable methods by the t-test with significance level $5\%$
are described in boldface.
]{
 \small{
 \begin{tabular}{@{} c |c c c c c c c @{}  }
   \multicolumn{8}{c}{}\\
 ID & RuLSIF & uLSIF & KLIEP & AR & SI & SST & OSVM \\ \hline
   1001 & .974 &.853 & .838 &.899 & .958 & .903 & .900 \\
   1002 & .996 &.963 & .909 &.872 & .969 & .880 & .905 \\
   1003 & .989 &.854 & .929 &.869 & .895 & .851 & .937 \\
   1004 & .996 &.868 & .890 &.881 & .941 & .886 & .891 \\
   1005 & .938 &.952 & .972 &.849 & .972 & .915 & .943 \\
   1006 & .933 &.918 & .889 &.778 & .890 & .925 & .842 \\
   1007 & .972 &.857 & .834 &.850 & .941 & .817 & .891 \\
   1008 & .995 &.922 & .930 &.892 & .981 & .860 & .907 \\
   1009 & .987 &.880 & .907 &.833 & .979 & .842 & .951 \\
   1010 & .991 &.952 & .889 &.821 & .915 & .867 & .903 \\
   \hline
   Ave. & \textbf{.977} &.902 & .900& .854 & .944 & .875 & .907 \\
   Std. & \textbf{.024} & .044 & .042 & .037 & .034& .034 & .032 \\
   \multicolumn{8}{c}{}
\end{tabular}
\label{tab.exp.human}
}}
\caption{HASC human-activity dataset.}
\label{fig.human}
\end{minipage}
\end{figure*}

In Figure~\ref{fig.illus.human}, examples of original time-series, true change points,
and change-point scores obtained by the RuLSIF-based method are plotted.
This shows
that the change-point score clearly captures
trends of changing behaviors,
except the changes around time 1200 and 1500.
However, because these changes are difficult to be recognized even by human,
we do not regard them as critical flaws.
Figure~\ref{fig.roc.human} illustrates ROC curves averaged over $10$ datasets,
and Figure~\ref{tab.exp.human} describes AUC values
for each of the $10$ datasets.
The experimental results show that the proposed RuLSIF-based method
tends to perform better than other methods.


Next, we use the \emph{IPSJ SIG-SLP Corpora and Environments for Noisy Speech Recognition}
(CENSREC) dataset provided by National Institute of Informatics (NII)\footnote{\url{http://research.nii.ac.jp/src/eng/list/index.html}},
which records human voice in a noisy environment.
The task is to extract speech sections from recorded signals.
This dataset offers several voice recordings with different background noises
(e.g., noise of highway and restaurant).
Segmentation of the beginning and ending of human voice is manually annotated.
Note that we only use the annotations as the ground truth for the final performance evaluation,
not for change-point detection
(i.e., this experiment is still completely unsupervised).

\begin{figure*}
\begin{minipage}[t]{\textwidth}
\centering
\subfigure[One of the original signals and change-point scores obtained by the RuLSIF-based method]{
 \includegraphics[width=.8\textwidth,clip]{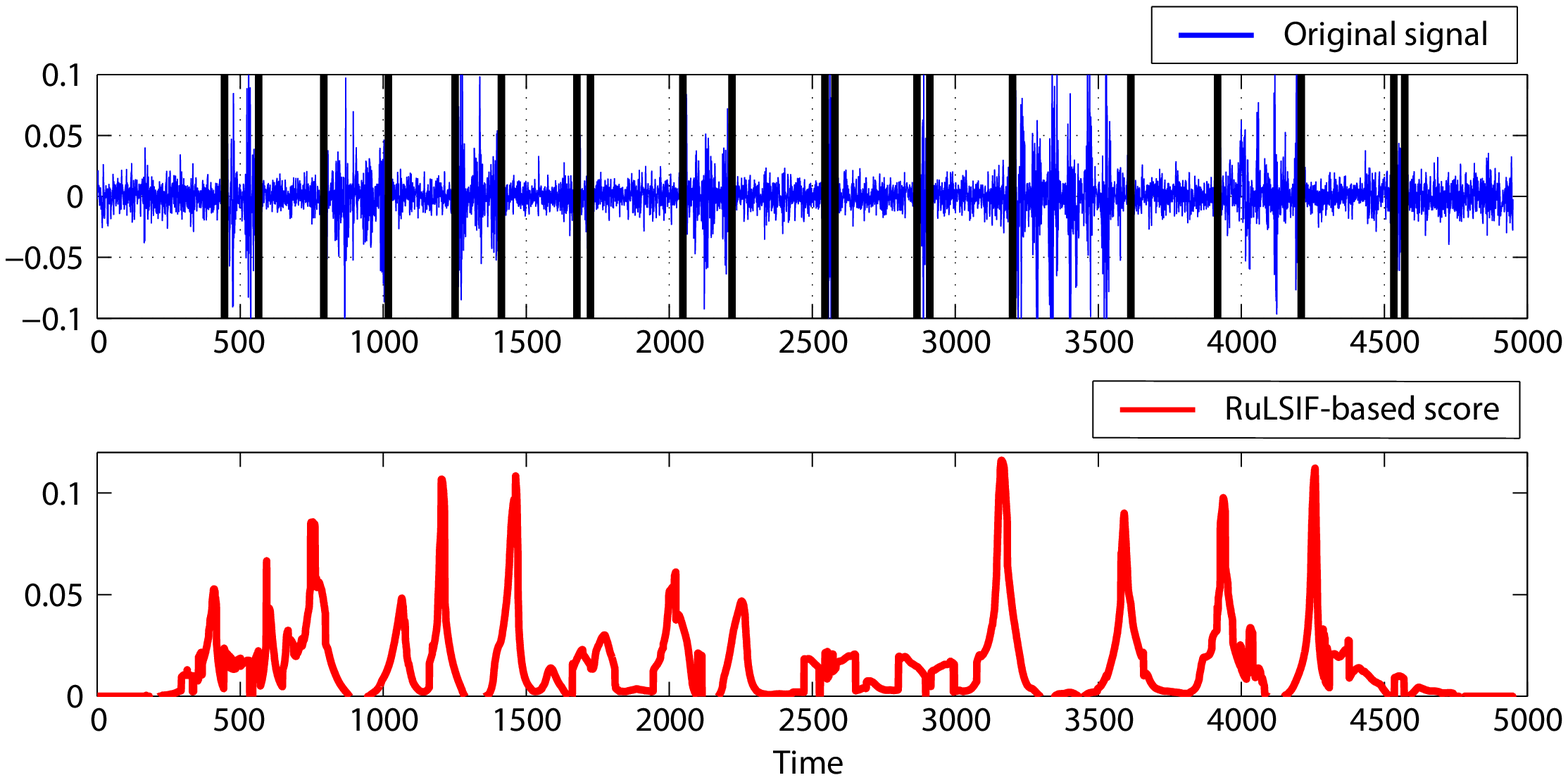}
 \label{fig.illus.speech}
 }
 \centering
 \subfigure[Average ROC curves]{
 \includegraphics[width=.5\textwidth,clip]{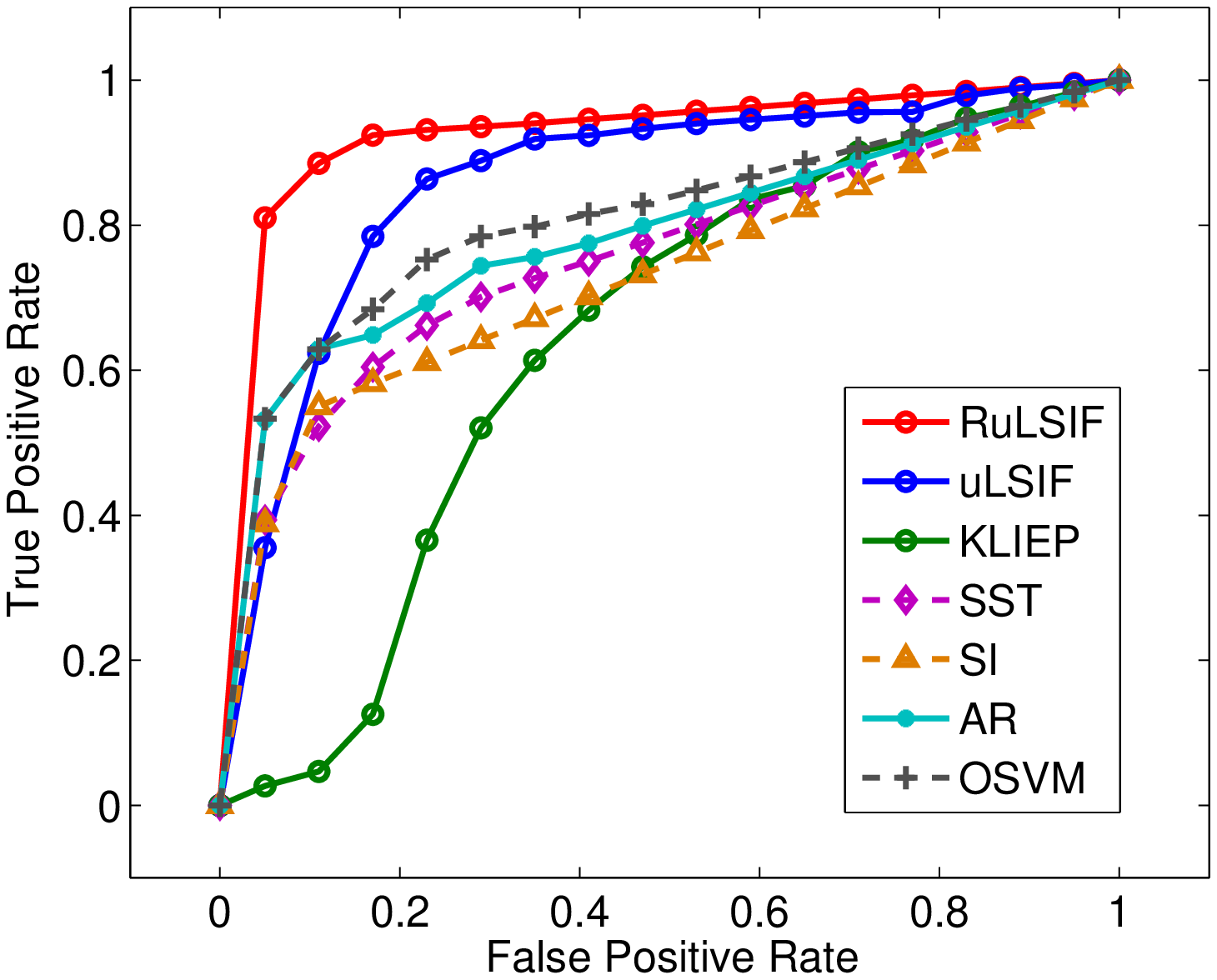}
 \label{fig.roc.speech}
 }
 \centering
 \subfigure[AUC values. The best and comparable methods by the t-test with significance level $5\%$
are described in boldface.]{\small{
 \begin{tabular}{@{} c |c c c c c c c @{}  }
   \multicolumn{8}{c}{}\\
 ID & RuLSIF & uLSIF & KLIEP & AR & SI & SST & OSVM \\ \hline
  01 & 1.00 & .902& .650& .860 & .690 & .806 & .800\\
  02 & .911 & .845& .712& .733 & .800 & .745 & .725\\
  03 & .963 & .931& .708& .910 & .899 & .807 & .932\\
  04 & .903 & .813& .587& .816 & .735 & .685 & .751\\
  05 & .927 & .907& .565& .831 & .823 & .809 & .840\\
  06 & .857 & .913 & .676& .868 & .740 & .736 & .838\\
  07 & .987 & .797& .657& .807 & .759 & .797 & .829\\
  08 & .962 & .757& .581& .629 & .704 & .682 & .800\\
  09 & .924 & .913& .693& .738 & .744 & .781 & .790\\
  10 & .966 & .856& .554&  .796 & .725 & .790 & .850\\
  \hline
   Ave. & \textbf{.940} & .863& .638& .798 & .762 & .764 & .815 \\
   Std. & \textbf{.044} & .059 & .061 &.081 & .063 & .049 & .057\\
   \multicolumn{8}{c}{}\\
\end{tabular}
\label{tab.exp.speech}
}}
\caption{CENSREC speech dataset.}
\label{fig.speech}
\end{minipage}
\end{figure*}

Figure~\ref{fig.illus.speech} illustrates an example of the original signals,
true change-points,
and change-point scores obtained by the proposed RuLSIF-based method.
This shows that the proposed method still gives clear indications for speech segments.
Figure~\ref{fig.roc.speech} and Figure~\ref{tab.exp.speech}
show average ROC curves over $10$ datasets and AUC values
for each of the $10$ datasets.
The results show that the proposed method significantly
outperforms other methods.



\subsection{Twitter Dataset}
\label{sec:twitter}
Finally, we apply the proposed change-point detection method to the
\emph{CMU Twitter dataset}\footnote{\url{http://www.ark.cs.cmu.edu/tweets/}},
which is an archive of Twitter messages 
collected from February 2010 to October 2010 via the Twitter application programming interface.


\begin{figure*}[t]
\centering
\vspace*{-8mm}
\subfigure[Normalized frequencies of 10 keywords]{
  \includegraphics[width=1\textwidth,clip]{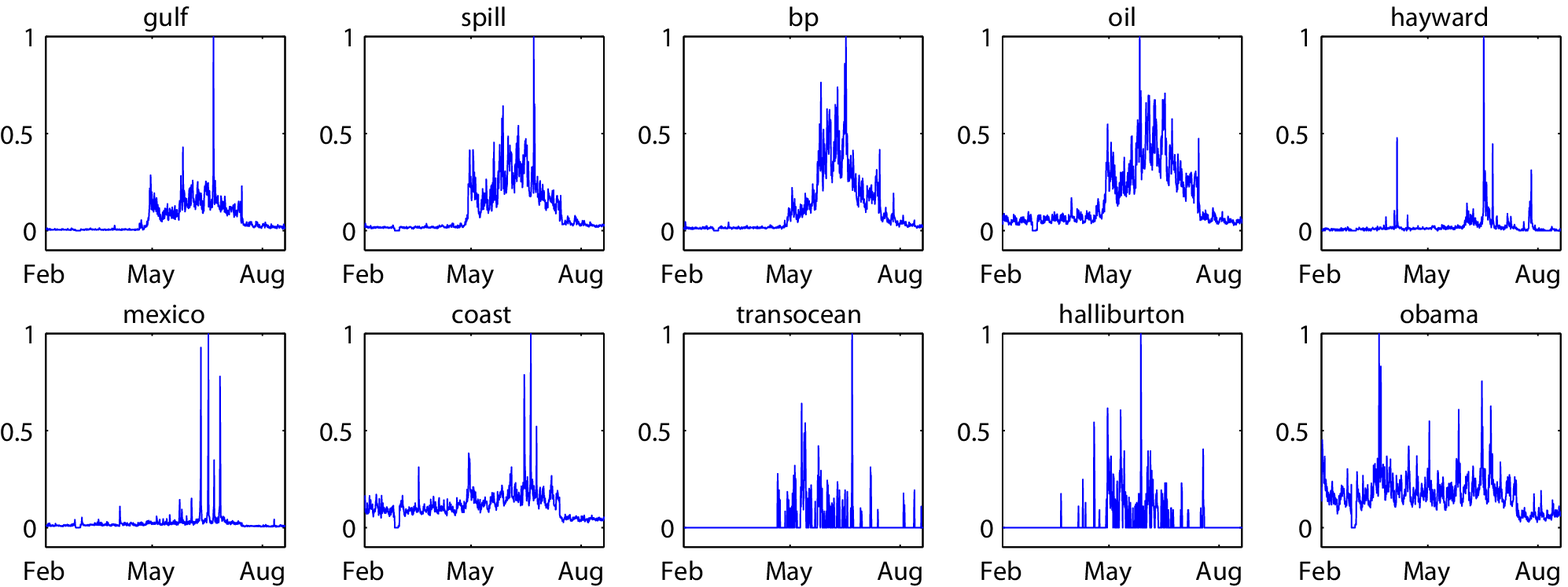}
 \label{fig.illus.twitter.data}
 }
\subfigure[Change-point score obtained by the RuLSIF-based method and exemplary real-world events]{
  \includegraphics[width=1\textwidth,clip]{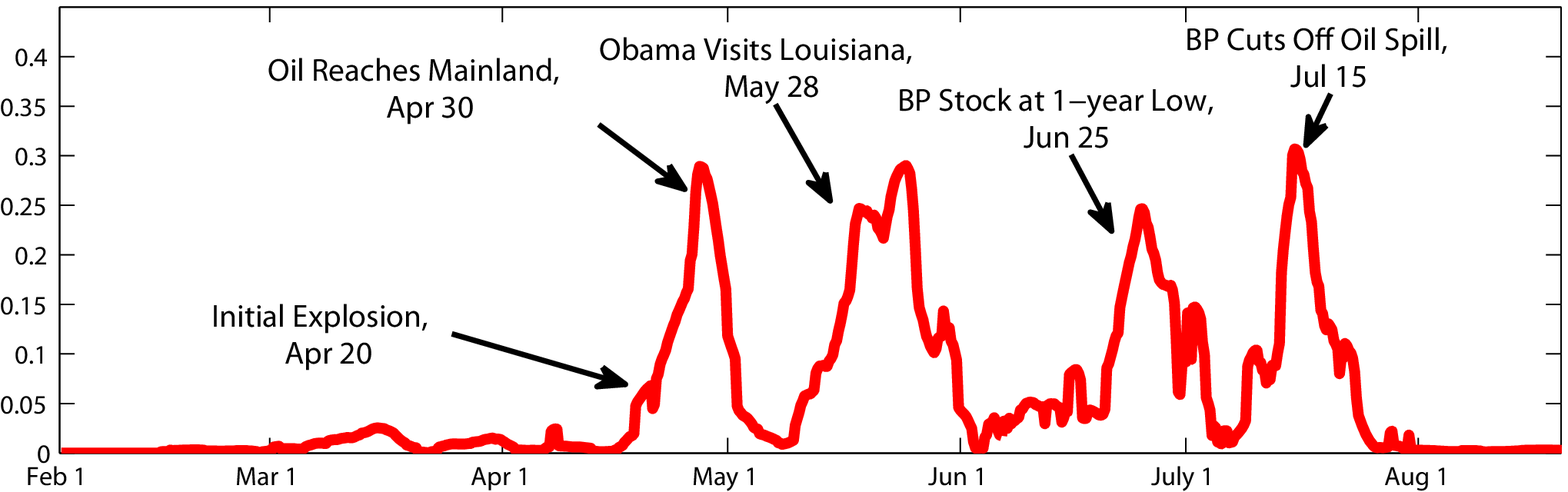}
 \label{fig.illus.twitter.CPDT}
 }
\caption{Twitter dataset.}
\label{fig.illus.twitter}
\end{figure*}

Here we track the degree of popularity of a given topic
by monitoring the frequency of selected keywords.
More specifically, we focus on events related to
``\emph{Deepwater Horizon oil spill in the Gulf of Mexico}''
which occurred on April 20, 2010\footnote{
\url{http://en.wikipedia.org/wiki/Deepwater_Horizon_oil_spill}}, and was widely broadcast among the Twitter community.
We use the frequencies of 10 keywords:
``\emph{gulf}'', ``\emph{spill}'', ``\emph{bp}'', ``\emph{oil}'',
``\emph{hayward}'', ``\emph{mexico}'', ``\emph{coast}'', ``\emph{transocean}'',  ``\emph{halliburton}'', and ``\emph{obama}''
(see Figure~\ref{fig.illus.twitter.data}).
We perform change-point detection directly on the 10-dimensional data,
with the hope that we can capture correlation changes between multiple keywords,
in addition to changes in the frequency of each keyword.


For quantitative evaluation, we referred to the Wikipedia entry ``Timeline of the Deepwater Horizon oil spill''\footnote{\url{http://en.wikipedia.org/wiki/Timeline_of_the_Deepwater_Horizon_oil_spill}} as a real-world event source. The change-point score obtained by the proposed RuLSIF-based method is plotted in Figure~\ref{fig.illus.twitter.CPDT}, where four occurrences of important real-world events show the development of this news story.

As we can see from Figure~\ref{fig.illus.twitter.CPDT}, the change-point score increases immediately after the initial explosion of the deepwater horizon oil platform and soon reaches the first peak when oil was found on the sea shore of Louisiana on April 30. Shortly after BP announced its preliminary estimation on the amount of leaking oil, the change-point score rises quickly again and reaches its second peak at the end of May, at which time President Obama visited Louisiana to assure local residents of the federal government's support.
On June 25, the BP stock was at its one year's lowest price, while the change-point score spikes at the third time. Finally, BP cut off the spill on July 15, as the score reaches its last peak.



\section{Conclusion and Future Perspectives}\label{sec:conclusion}
In this paper, we first formulated the problem of retrospective change-point detection
as the problem of comparing two probability distributions over two consecutive time segments.
We then provided a comprehensive review of state-of-the-art
density-ratio and divergence estimation methods,
which are key building blocks of our change-point detection methods.
Our contributions in this paper were
to extend the existing KLIEP-based change-point detection method \citep{CPDT_JOUR},
and to propose to use uLSIF as a building block.
uLSIF has various theoretical and practical advantages, for example,
the uLSIF solution can be computed analytically,
it possesses the optimal non-parametric convergence rate,
it has the optimal numerical stability, and it has higher robustness than KLIEP.
We further proposed to use RuLSIF, a novel divergence estimation paradigm
emerged in the machine learning community recently.
RuLSIF inherits good properties of uLSIF, and moreover it possesses
an even better non-parametric convergence property.
Through extensive experiments on artificial
datasets and real-world datasets
including human-activity sensing, speech, and Twitter messages,
we demonstrated that the proposed RuLSIF-based change-point detection
method is promising.

Though we estimated a density ratio between two consecutive segments, some earlier researches \citep{Detection_Abrupt_Changes,GenLR,book:Gustafsson:2000} introduced a hyper-parameter that controls the size of a margin between two segments. In our preliminary experiments, however, we did not observe significant improvement by changing the margin. For this reason, we decided to use a straightforward model that two segments have no margin in between.


Through the experiment illustrated in Figure \ref{fig.illus.n.k.AUC} in Section \ref{sec.artifi}, we can see that the performance of the proposed method is affected by the choice of hyper-parameters $n$ and $k$. However, discovering optimal values for these parameters remains a challenge, which will be investigated in our future work.

RuLSIF was shown to possess a better convergence property than uLSIF
\citep{RULSIF_NIPS} in terms of density ratio estimation.
However, how this theoretical advantage in density ratio estimation
can be translated into practical performance improvement in change detection is still not clear, beyond the intuition that a better divergence estimator gives a better change score. We will address this issue more formally in the future work.


Although the proposed RuLSIF-based change-point detection was shown to work well
even for multi-dimensional time-series data,
its accuracy may be further improved by incorporating \emph{dimensionality reduction}.
Recently, several attempts were made to combine dimensionality reduction
with direct density-ratio estimation
\citep{NN:Sugiyama+etal:2010,NN:Sugiyama+etal:2011a,AAAI:Yamada+Sugiyama:2011}.
Our future work will apply these techniques to change-point detection
and evaluate their practical usefulness.

Compared with other approaches,
methods based on density ratio estimation tend to be computationally more expensive
because of the cross-validation procedure for model selection.
However, thanks to the analytic solution, 
the RuLSIF- and uLSIF-based methods are computationally more efficient than
the KLIEP-based method that requires an iterative optimization procedure (see Figure 9 in \citet{ULSIF} for the detailed time comparison between uLSIF and KLIEP).
Our important future work is to further improve the computational efficiency of
the RuLSIF-based method.

In this paper, we focused on computing the change-point score
that represents the plausibility of change points.
Another possible formulation is hypothesis testing,
which provides a useful threshold to determine whether a point is a change point.
Methodologically, it is  straightforward to extend the proposed method
to produce the $p$-values, following the recent literatures
\citep{NN:Sugiyama+etal:2011b,IEEE-IT:Kanamori+etal:2011}.
However, computing the $p$-value is often time consuming,
particularly in a non-parametric setup.
Thus, overcoming the computational bottleneck is an important future work
for making this approach more practical.

Recent reports pointed out that Twitter messages can be indicative of real-world events \citep{StreamingStory2010,EarthquakeTwitter2010}.
Following this line,
we showed in Section~\ref{sec:twitter} that
our change-detection method can be used as a novel tool
for analyzing Twitter messages.
An important future challenge along this line
includes automatic keyword selection for topics of interests.



\section*{Acknowledgements}
SL was supported by NII internship fund and the JST PRESTO program.
MY and MS were supported by the JST PRESTO program.
NC was supported by NII Grand Challenge project fund.
\bibliography{main_sspr}

\begin{thebibliography}{54}
\bibitem[Pearson(1900)]{PE_DIV}
K.~Pearson.
\newblock On the criterion that a given system of deviations from the probable
  in the case of a correlated system of variables is such that it can be
  reasonably supposed to have arisen from random sampling.
\newblock \emph{Philosophical Magazine}, 50:\penalty0 157--175, 1900.
\end{thebibliography}

\end{document}